\documentclass[11pt,noindent]{article}
\usepackage{latexsym,amsmath,amsfonts}
\usepackage{xcolor}
\usepackage{amsmath,amssymb,epsfig}
\usepackage[shortcuts]{extdash}
\usepackage{authblk}
\usepackage[symbol]{footmisc}

\usepackage[T1]{fontenc}
\usepackage[margin=3cm]{geometry}
\usepackage[utf8]{inputenc}
\usepackage{array}
\usepackage[colorlinks=true,citecolor=blue,linkcolor=blue,urlcolor=blue]{hyperref}
\usepackage[backend=biber,date=year,
           sortcites=true,defernumbers=true,
           maxbibnames=99,giveninits=true,uniquename=false,url=false,bibstyle=numeric]{biblatex}
\newcommand{\HQ}{H²Q}
\newcommand{\Hyp}{HyP²}

\newcommand{\appendixhead}%
{\vspace{0.25in}\noindent\textbf{\huge Appendices}}

\usepackage{cleveref}
\usepackage{setspace}
\usepackage{changepage}
\usepackage[shortlabels]{enumitem}
\allowdisplaybreaks
\usepackage{caption}
\usepackage{subcaption}

\usepackage[noline, boxed]{algorithm2e}
\SetKwInOut{Parameter}{parameter}

\makeatletter
\let\blx@rerun@biber\relax
\makeatother

\newtheorem{theorem}{Theorem}

\numberwithin{theorem}{section} \numberwithin{equation}{section}
\numberwithin{proposition}{section} \numberwithin{lemma}{section}
\numberwithin{corollary}{section}

\newcommand\ignore[1]{}

\def\R{\mathbb{R}} 

\newcommand\QED{\ifhmode\allowbreak\else\nobreak\fi
\quad\nobreak$\Box$\medbreak}
\newcommand{\proofstart}{\par\noindent\sl Proof:\rm\enspace}
\newcommand{\proofend}{\QED\par}
\newenvironment{proof}{\proofstart}{\proofend}

\newcommand{\eg}{e.g.,~}
\newcommand{\ie}{i.e.,~}

\DeclareMathOperator*{\argmin}{arg\,min}
\DeclareMathOperator*{\sign}{sign}

\title{Deep Hashing via Householder Quantization}
\stepcounter{footnote}
\author[1]{Lucas R. Schwengber\footnote{These authors contributed equally.}}
\author[1]{Lucas Resende\textsuperscript{†}}
\author[1]{Paulo Orenstein}
\author[1]{Roberto I. Oliveira}
\affil[1]{IMPA, Rio de Janeiro, Brazil}

\begin{document}

\maketitle

\begin{abstract}
    Hashing is at the heart of large-scale image similarity search, and recent methods have been substantially improved through deep learning techniques. Such algorithms typically learn continuous embeddings of the data. To avoid a subsequent costly binarization step, a common solution is to employ loss functions that combine a similarity learning term (to ensure similar images are grouped to nearby embeddings) and a quantization penalty term (to ensure that the embedding entries are close to binarized entries, \eg, -1 or 1). Still, the interaction between these two terms can make learning harder and the embeddings worse. We propose an alternative quantization strategy that decomposes the learning problem in two stages: first, perform similarity learning over the embedding space with no quantization; second, find an optimal orthogonal transformation of the embeddings so each coordinate of the embedding is close to its sign, and then quantize the transformed embedding through the sign function. In the second step, we parametrize orthogonal transformations using Householder matrices to efficiently leverage stochastic gradient descent. Since similarity measures are usually invariant under orthogonal transformations,
    this quantization strategy comes at no cost in terms of performance. The resulting algorithm is unsupervised, fast, hyperparameter-free and can be run on top of any existing deep hashing or metric learning algorithm. We provide extensive experimental results showing that this approach leads to state-of-the-art performance on widely used image datasets, and, unlike other quantization strategies, brings consistent improvements in performance to existing deep hashing algorithms.
\end{abstract}

\section{Introduction}
\label{sec:intro}

With the massive growth of image databases \cite{deng2009imagenet, lin2014mscoco, chua2009nus, krizhevsky2009cifar10}, there has been an increasing need for fast image retrieval methods. Traditionally, hashing has been employed to quickly perform approximate nearest neighbor search while retaining good retrieval quality. For example, Locality-Sensitive Hashing (LSH) \cite{andoni2014lsh, gionis1999lsh, jafari2021lshsurvey} assigns compact binary hash codes to images such that similar items receive similar hash codes, so the search can be conducted over hashes rather than images. Still, LSH methods are agnostic to the nature of the underlying data, often leading to sub-optimal performance.

\begin{figure}[ht!]
  \centering
   \includegraphics[width=.5\linewidth]{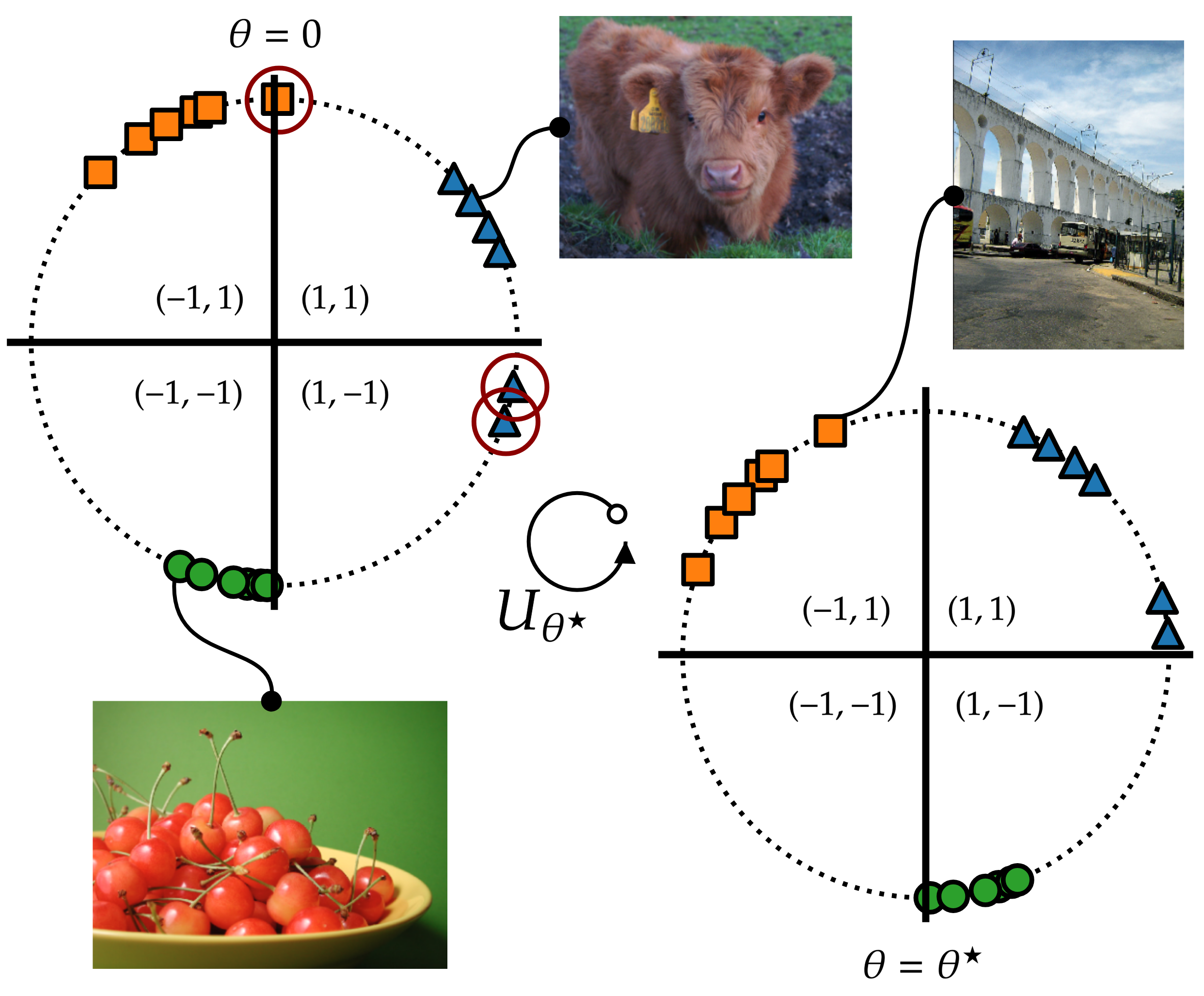}

   \caption{Deep hashing methods usually train low-dimensional embedding maps and obtain hashes by taking the sign of the embeddings coordinate-wise. To avoid a lossy discretization, a two-term loss $L = L_S + \lambda L_Q$ is used, where $L_S$ is a similarity term and $L_Q$ is a quantization term. In contrast, we first let $\lambda = 0$, obtaining good similarity-preserving embeddings and then train an orthogonal transformation $U_\theta$ parametrized by $\theta \in \Theta$ to binarize the embedding via the coordinate-wise sign. As similarity losses are typically invariant under $U_{\theta}$, the term $L_S$ remains unchanged as an optimal discretization is found through the choice of $\theta$.}
   \label{fig:onecol}
\end{figure}

Indeed, many data-aware methods have recently been proposed, giving rise to the learning to hash literature \cite{wang2017survey} and substantially improving the results for the image retrieval problem. In unsupervised learning to hash \cite{weiss2008spectral, wang2010semi, liu2011hashing, liu2014discrete, gong2012iterative}, one uses only feature information (\ie the image itself), while in supervised hashing \cite{shen2015discretesupervised, liu2012kernelsupervised} one also employs additional available label information (\eg a class the image belongs to) to capture semantic relationships between data points. With the significant advances brought about by deep learning in image tasks, deep hashing methods \cite{luo2023dlsurvey, singh2022dlsurvey} have become state-of-the-art methods in the field. 

While hashing is discrete in nature, optimizing directly over binary hashes is usually intractable \cite{wang2015learning}. To overcome this, one alternative is to simply learn efficient embeddings through a similarity-learning term in the loss function, and then binarize the embedding (typically by using the coordinate-wise sign of the embedding, so they become either $-1$ or $1$). However, this quantization process can significantly decrease the retrieval quality, so most deep hashing methods try to account for it in the learning process. That is, deep hashing methods look for solutions such that (i) the learned embeddings preserve similarity well, which is solved through a similarity learning strategy; (ii) the error between the embedding and its binarized version is small, which calls for a quantization strategy. Most deep hashing methods combine the similarity learning and quantization strategies in a single two-termed loss function where one term accounts for similarity learning and the other term penalizes quantization error. As it turns out, the experiments in this paper show that learning similarity and quantization at the same time is often sub-optimal.

We propose to decouple the problem in two independent stages. First, we minimize the similarity term of the deep hashing method (\eg, \cite{hyp22022, hashnet2017, dch2018, dhn2016, dpsh2015}) with no quantization penalty to obtain a continuous embedding for the images. Then, our quantization strategy consists in training a rotation (or, more generally an orthogonal transformation) that when applied to the previously obtained embeddings minimizes the distance between the embeddings and their binary discretization. Because similarity learning losses in deep hashing are invariant under rotations, this quantization strategy is effectively independent from the similarity step, so each step can be solved optimally.

More broadly, the process of optimizing an orthogonal transformation to reduce lossy compression from the quantization may be applied to any pre-trained embedding. We provide comprehensive experimental results to show that this two-step learning process significantly increases the performance metrics for most deep hashing methods, beating current state-of-the-art solutions, at a very low computation cost and with no hyperparameter tuning. In summary, our main contributions are:
\begin{itemize}
    \item We propose a new quantization method called Householder hashing quantization (\HQ), which turns pre-trained embeddings into efficient hashes. The hash is created in two steps: (i) finding a good embedding of the data through some similarity learning strategy, and (ii) quantizing it after using optimal orthogonal transformations via Householder transforms (see \Cref{fig:onecol}).
    \item While existing deep hashing methods often combine the two steps through a quantization term in the loss function, the strategy above typically yields better results by exploring an invariance in the similarity term to orthogonal transformations (see \Cref{sec:H2Q}). Thus, for state-of-the-art hashing methods such as \Hyp, Householder quantization uniformly improves performance; this is not the case for other quantization strategies (see \Cref{sec:uniform_improvements_on_similarity_losses,sec:other_quantizations}).
    \item Our algorithm is unsupervised, fast and linear in the size of the data (\Cref{sec:computational_cost}). In contrast to most current quantization strategies, our method does not require hyperparameters. It can also be run atop any existing deep hashing or metric learning algorithm. 
    \item In several experiments with NUS WIDE, MS COCO, CIFAR 10 and ImageNet datasets, we show that Householder quantization significantly helps the best performing hashing methods in the literature, delivering state-of-the-art results versus current benchmarks (\Cref{sec:improve_sota}).
\end{itemize}

\section{Related Work}

There is a vast literature on unsupervised hashing methods for image retrieval. An important early example is Locality-Sensitive Hashing \cite{gionis1999lsh, andoni2014lsh, jafari2021lshsurvey}, a data-agnostic framework that builds random hash functions such that similar images are mapped to similar hashes and so retrieval achieves sub-linear time complexity. Still, it is usually possible to build better hashes by learning the hash functions from the data under consideration, so many learning to hash methods have been proposed. For example, Spectral Hashing \cite{weiss2008spectral} and Semi-Supervised Hashing (SSH) \cite{wang2010semi} build on principal component analysis (PCA) to create data-aware embeddings which are then binarized using the sign function. 

More recently, deep hashing methods have significantly advanced the state-of-the-art results for fast image retrieval. These methods compose the last layer of pre-trained convolutional neural network (CNN) architectures, such as AlexNet \cite{krizhevsky2012imagenet}, VGG-11 and VGG-16 \cite{simonyan2014very}, with a sequence of fully connected layers to be fine-tuned. By using pre-trained architectures, they exploit the enriched features and start the training procedure with an embedding that already encodes a high level of semantic similarity between images. Convolutional Neural Network Hashing (CNNH) \cite{xia2014cnnh} was one of the first methods of this type; it first finds a binary encoding that approximates the similarity between data points, and then trains a CNN to map the original data points into this binary encoding. Deep Supervised Hashing (DSH) \cite{dsh2016} considers a loss function with a similarity term which is analogous to the contrastive squared losses used in metric learning \cite{kulis2013metricsurvey, chopra2005metriclearning} while adding a penalization term in terms of the $L_1$ loss. On the other hand, Deep Hashing Network (DHN) \cite{dhn2016} considers a pairwise cross-entropy loss for the similarity term, while using the same $L_1$ quantization term. HashNet \cite{hashnet2017} builds on DHN by adding weights to counter the imbalance between the number of positive and negative paris, and also applies a hyperbolic tangent to the embedding to continuously approximate the sign function used in the binarization step. Deep Cauchy Hashing \cite{dch2018} and alternatives \cite{weighted2021, kang2019mmhh, li2015dpsh, cao2018dph} follow a similar strategy with variations on the choice of the similarity and penalization terms and the weights. Methods such as Pairwise Correlation Discrete Hashing (PCDH) \cite{chen2020pcdh} and Deep Supervised Discrete Hashing (DSDH) \cite{li2017dsdh} additionally consider how well the hash codes can reconstruct available labels by using a classifier. An alternative to training with pairwise similarity losses is to use a triplet loss, as is the case for Deep Neural Networks Hashing (DNNH) \cite{lai2015triplet}. Another alternative are proxy-based methods \cite{hoe2021one, yuan2020central, fan2020deepolarized} such as OrthoHash \cite{hoe2021one} which maximizes the cosine similarity between data points and pre-defined target hash codes associated with each class. Fixing target hash codes might miss semantic relationships between class labels, so \cite{aziere2019ensemble, kim2020proxy} consider the hash centers as parameters to be learned. More recently, \Hyp \cite{xu2022hyp2} combined a proxy-based loss with a pairwise similarity term, harnessing the power from both approaches to obtain state-of-the-art performance. This suggests that reducing quantization error through a penalty term is not a requirement for good performance in deep hashing methods.

Indeed, we build on these latest deep hashing methods by introducing a novel quantization strategy that exploits the gains in similarity learning at no cost in terms of quantization. It consists of efficiently binarizing the learned embeddings after applying an optimal orthogonal transformations obtained via stochastic gradient descent \cite{fasth}, based on a parametrization using Householder matrices. The orthogonal transformation is optimized to make the embedding entries as close to $-1, 1$ as possible before the coordinate-wise sign function is applied. This approach is similar in spirit to other quantization strategies developed before deep hashing (\eg, Iterative Quantization \cite{gong2012iterative} and similar methods \cite{jegou2008hamming, jegou2010aggregating, wang2015learning}), but with important differences. In \cite{jegou2008hamming}, the authors propose using random orthogonal transformations to improve hash codes, and, in \cite{jegou2010aggregating}, the authors quantize database vectors with centroids using orthogonal Householder reflections. Both \cite{gong2012iterative} and \cite{wang2015learning} propose iterative algorithms to learn orthogonal transformations under an $L_2$ loss. They iteratively solve an orthogonal Procrustes problem for fixed hash codes, and then find the binarized hash code given the orthogonal transformation found. More recently, HWSD \cite{doan2022one} replaces each deep hashing method's penalty term by a sliced Wasserstein distance. Our method differs from earlier rotation-based schemes \cite{gong2012iterative, jegou2008hamming, jegou2010aggregating, wang2015learning} since it is not iterative and exploits the capabilities of SGD to avoid spurious local minima; it also differs from HWSD because we do not jointly optimize embeddings and quantization. Furthermore, while previous quantization strategies sometimes degrade the embeddings learned by deep hashing algorithms, our proposed quantization strategy uniformly improves them (see \Cref{tab:sota_CNNF_alexnet,tab:improve_CNNF_alexnet}, and \Cref{fig:other_quantizations}).


\section{Householder Hashing Quantization (\HQ)}
\label{sec:H2Q}

In learning to hash, we are given a set of images $\{ x_i \}_{i=1}^n$, $x_i \in \mathbb{R}^d$, and a notion of similarity between pairs of images $\mathcal{S} = (s_{ij})_{i,j \in [n]}$ taken to be $1$ for similar images (\eg, from the same object or class) and $0$ for dissimilar ones. 
The goal is to learn a hash function $h_{\theta}: \R^{d} \mapsto \{ -1, 1 \}^k$, with associated hash codes $b_i = h_{\theta}(x_i)$, such that the Hamming distance $d_H(b_i,b_j)$ between $b_i$ and $b_j$,
\begin{equation*}
  d_H(b_i,b_j) = \sum_{l=1}^k \textbf{1}_{[b_{il} \neq b_{jl}]},
\end{equation*}
is small for similar pairs (\ie, $s_{ij}=1$) and big for dissimilar ones ($s_{ij}=0$). Since optimizing over $b_{i}\in\{-1,1\}^{k}$ is computationally intractable \cite{wang2015learning}, one usually learns a continuous embedding $f_{\theta}:\R^{d} \to \R^{k}$, with $\theta \in \Theta$ a parameter to be learned, and then binarize it via $h_{\theta} = \sign \circ f_{\theta}$.

\subsection{Deep Hashing Losses}

In deep hashing, $\theta$ represents the weights of a neural network $f_{\theta}$, to be learned by stochastic gradient descent (SGD) using some loss function. Generally, one would like to directly optimize them in terms of $d_{ij}(\theta) = d_H( h_\theta(x_i) , h_\theta(x_j) )$, solving the minimization problem
\begin{equation}
\label{eq:ideal_loss}
    \min_{\theta} \sum_{i,j} s_{ij} l_S( d_{ij}(\theta) ) + (1-s_{ij})l_D( d_{ij}(\theta) ),
\end{equation}
where $l_S$ and $l_D$ are losses for similar and dissimilar pairs of points, respectively. Many alternatives for $l_S$ and $l_D$ have been proposed (\eg, \cite{luo2023dlsurvey, wang2017survey, hashnet2017, dch2018, dpsh2015, dhn2016, weighted2021, hyp22022}).

However, due to the discrete nature of hash functions, the objective function in \eqref{eq:ideal_loss} is not differentiable in $\theta$. A common way to overcome this is to consider the identity: 
\begin{equation}
\label{eq:cos_approx}
\begin{split}
    d_H(b_i,b_j) &= \frac{k -\langle b_i, b_j \rangle}{2} =  \frac{k}{2} \left(1 - \frac{\langle b_i, b_j \rangle}{\| b_i \|_2 \| b_j \|_2} \right),
\end{split}
\end{equation}
which holds when $b_i, b_j \in \{-1,1\}^k$ are hash codes. This identity relates the Hamming distance with the inner product and the cosine similarity. By replacing $b_i$ and $b_j$ with $f_i = f_{\theta}(x_i)$ and $f_j = f_{\theta}(x_j)$ the last two expressions in \eqref{eq:cos_approx} yield ways to measure the distance between pairs of points in the embedding, generalizing the Hamming distance to a differentiable expression. Thus, letting $\tilde{d}_{ij}(\theta)$ be either $(k -\langle f_i, f_j \rangle)/2$ or $(k/2) \left(1 - \langle f_i, f_j \rangle/(\| f_i \|_2 \| f_j \|_2) \right)$, a relaxed version of \eqref{eq:ideal_loss} consists in minimizing the objective
\begin{equation}
\label{eq:gen_sim_loss_func}
    L_S(\theta) = \sum_{i,j} s_{ij} l_S\left( \Tilde{d}_{ij}(\theta) \right) + (1-s_{ij}) l_D\left( \Tilde{d}_{ij}(\theta) \right).
\end{equation}

Once the embedding $f_{\theta}$ is learned, one typically binarizes it by taking its coordinate-wise sign (\ie, $b(x_{i}) = \sign \circ f_{\theta}(x_i)$), but this can be quite lossy when $f_{\theta}(x_i)$ is very far from $\{ -1, 1 \}^k$. To mitigate this effect, a penalty term is usually introduced to minimize the gap between $f_i$ and $b_i$. This gives rise to the two-term loss function:
\begin{equation}
\label{eq:two_terms}
    L(\theta) = L_S(\theta) + \lambda \cdot L_Q(\theta),
\end{equation}
where $L_S(\theta)$ takes the form in \eqref{eq:gen_sim_loss_func}, $\lambda \in \R$ and a typical example of $L_Q(\theta)$ (\eg, used in \cite{dpsh2015, dsh2016, zhang2016discrete, weighted2021}) would be:
\begin{equation}
\label{eq:quant_term_l2}
    L_Q(\theta) = \frac{1}{n}\sum_{i=1}^n \| f_\theta(x_i) - h_\theta(x_i) \|_2^2,
\end{equation}
Other quantization losses are explored in \cite{dhn2016, dch2018}. Note, however, that this quantization strategy directly affects the learned embedding $f_{\theta}$ since (\ref{eq:two_terms}) trains on both $L_S(\theta)$ and $\lambda \cdot L_Q(\theta)$, resulting in possibly subpar hashing performance.

\subsection{The H²Q Quantization Procedure}

We propose instead to decompose the similarity learning and the quantization strategies in two separate steps. First, for a given deep hashing method, set $\lambda = 0$ in \eqref{eq:two_terms} and solve
\begin{equation} 
\label{eq:sim_based_loss}
    \min_{\theta \in \Theta} L_S(\theta)
\end{equation}
to learn an embedding $f_\theta$ that preserves similarity as well as possible. 
Then, normalize $f_\theta(x_i)$ to obtain
\[\Bar{f}_\theta(x_i) = \sqrt{k}\frac{f_\theta(x_i)}{\| f_\theta(x_i) \|_2}\]
and, finally, solve the following optimization problem:
\begin{equation}
\label{eq:orthogonal_transformation_optimization}
    U^{\star} = \argmin_{U \in O(k)}\frac{1}{n}\sum_{i=1}^n \left\| U \Bar{f}_\theta(x_i) - \sign\left( U  \Bar{f}_\theta(x_i) \right) \right\|_2^2.
\end{equation}
 Note this is minimizing a quantization error over the group $O(k)$ of orthogonal transformations in $\R^k$. We then obtain the final \HQ-quantized hash function $h_{\theta} = \sign(U^{\star} f_{\theta})$.

\RestyleAlgo{boxruled}
\begin{algorithm}[ht!]
  \caption{\HQ\ quantization strategy. \label{alg:h2q}}
  
  \textbf{Input:} embeddings $f_1, \dots, f_n \in \R^k$ of images $x_1, \dots, x_n$ trained with a similarity-based loss \eqref{eq:sim_based_loss}.


  \textbf{Procedure}
  \begin{enumerate}
      \item Compute $\Bar{f_i} = \sqrt{k}\frac{f_i}{\| f_i \|_2}$ for $i=1, \ldots, n$;
      \item Solve, using SGD,
      \[ U^\star = \argmin_{U \in O(k) } \frac{1}{n} \sum_{i=1}^n \left\| U\Bar{f_i} - \sign\left(U \Bar{f}_i\right) \right\|_2^2, \]
      where $O(k)$ is the orthogonal group parametrized\\via Householder matrices (see Section \ref{sec:parametrization});
      \item Evaluate $h_i = \sign(U^\star f_i) \in \{-1, 1\}^{k}$, $i=1, \ldots, n$;
      \item Output hashes $h_1, \ldots, h_n$ for images $x_{1}, \ldots, x_{n}$.
  \end{enumerate}
\end{algorithm}

Our proposed procedure is summarized in Algorithm \ref{alg:h2q}. The normalization is required to avoid excessive penalization towards embeddings $f_i$ with larger norms $\| f_i \|_2$. Still, we note that, once $U^{\star}$ is found, the prediction can be done without normalization. The factor of $\sqrt{k}$ puts the normalized features in the Euclidean sphere containing the hash codes. Finally, the reason we use orthogonal transformations is due to the following result from linear algebra:
\begin{theorem}
\label{thm:inner_prod}
    A map $U: \mathbb{R}^k \mapsto \mathbb{R}^k$ preserves inner products if and only if it is a linear orthogonal transformation.
\end{theorem}
\begin{proof}
    See \Cref{sm:proofs} in the Supplement.
\end{proof}

Since cosine similarity depends only on inner products, it follows that orthogonal transformations also preserve cosine similarity between points. Thus, we are effectively optimizing our quantization strategy over the largest possible set of transformations that make the term $L_S$ invariant for the deep hashing. Hence, the quality of the embedding is not sacrificed due to the subsequent quantization. 

\subsection{Parametrizing Orthogonal Transformations}\label{sec:parametrization}

Finding the right parametrization of the orthogonal group $O(k)$ to solve \eqref{eq:orthogonal_transformation_optimization} is non-trivial. One may consider, for example, matrix exponentials or Cayley maps \cite{lezcano2019trivializations, absil2009optimization, golub1996matrix, li2020cayley}. We propose to parametrize the elements of the group $O(k)$ as the product of Householder matrices. Geometrically, a Householder matrix is a reflection about a hyperplane with normal vector $v \in \R^k \setminus \{0\}$ and containing the origin, \ie,
\begin{equation}
 H = I_{k} - 2 \frac{v v^\intercal}{\| v \|_2^2},
\end{equation}
where $I_k \in \R^{k \times k}$ is the identity. Any orthogonal matrix can be decomposed as a product of Householder matrices:
\begin{theorem}
\label{thm:householder_decomp}
    For every orthogonal matrix $U \in O(k)$, there exists vectors $v_1, \dots, v_k \in \R^k\setminus\{0\}$ such that $U$ is the composition of their respective Householder matrices, \ie,
    \begin{equation}
    \label{eq:householder_decomp}
    U = \prod_{i=1}^k \left( I_k - 2 \frac{v_i v_i^\intercal}{\| v_i \|^2 } \right).
    \end{equation}
    Conversely, every matrix with the form \eqref{eq:householder_decomp} is orthogonal.
\end{theorem}
\begin{proof}
    See \Cref{sm:proofs} in the Supplement.
\end{proof}

Thus, finding an optimal orthogonal transformation is equivalent to finding optimal $v_1, \dots, v_k \in \R^k$, which can be thought of as parameters in \eqref{eq:orthogonal_transformation_optimization}, and learned through SGD. Since solving it using SGD requires several batch-evaluations of $U \Bar{f_i}$, we use the matrix multiplication algorithm in \cite{fasth} to perform this operation efficiently.

\section{Experiments} \label{sec:experiments}

In this section we present experimental evidence sustaining the following three claims:
\begin{enumerate}
    \item \HQ\ is capable of improving state-of-the-art hashing, obtaining the best performance metrics over existing deep hashing alternatives;

    \item \HQ\ always improves the metrics of cosine and inner-product similarity-based losses;

    \item In contrast, other quantization strategies, such as ITQ \cite{gong2012iterative}, HWSD \cite{doan2022one} and penalization terms (\eg, (\ref{eq:quant_term_l2})), may deteriorate performance, sometimes significantly.
\end{enumerate}
Moreover, we also provide experiments regarding computational time and an ablation study where variants of \eqref{eq:orthogonal_transformation_optimization} are discussed. We start by introducing our experimental setup.

\vspace{1em}

\textbf{Datasets.} We consider four popular image retrieval datasets, of varying sizes: CIFAR 10 \cite{cifar10}, NUS WIDE \cite{nuswide}, MS COCO \cite{mscoco} and ImageNet \cite{imagenet}.

\textit{CIFAR 10} \cite{cifar10} is an image dataset containing $60,000$ images divided into $10$ mutually exclusive classes. Following the literature \cite{dch2018,dhn2016}, we take $500$ images per class for the training set, $100$ images per class for the test and validation sets. The remaining images are used as database images.

\textit{NUS WIDE} \cite{nuswide} is a web image dataset containing a total of $269,648$ images from \href{https://www.flickr.com/}{flickr.com}. Each image contains annotations from a set of 81 possible concepts. Following \cite{hyp22022, zhang2020improved}, we first reduce the number of total images to $195,834$ by taking only images with at least one concept from the 21 most frequent ones. We then remove images that were unavailable for download from \href{https://www.flickr.com/}{flickr.com}, resulting in a set of $148,332$ images. From this subset we randomly sample $10,500$ images as training set and $2,100$ images for each of the query and validation sets. The remaining images compose our database, as in \cite{hyp22022, zhang2020improved}.

\textit{MS COCO} \cite{mscoco} is a dataset for image segmentation and captioning containing a total of $123,287$ images, $82,783$ from a training set and $40,500$ from a validation set. Each image has annotations from a list of $80$ semantic concepts. Following \cite{dch2018}, we randomly sample $10,000$ images as training set, $5,000$ images for each of the query and validation set, and the remaining images are used as database images.

\textit{ImageNet} \cite{imagenet} is a large image dataset with over $1,200,000$ images in the training set and $50,000$ in the validation set, each having a single label from a list of $1000$ possible categories. We use the same choice of $100$ categories as \cite{hashnet2017} resulting in the same training set, containing $13,000$ images, and database set, containing $128,503$ images. Finally, we split the $5,000$ images from the test set into $2,500$ images for each of the query and validation sets.

\vspace{1em}

\textbf{Evaluation metric.} The standard metric in learning to hash is mean average precision \cite{singh2022dlsurvey}. It measures not only the precision of the retrieved items, but also the ranking in which the items are retrieved. More precisely, let $q$ be a query image and $D = \{x_1, \dots, x_n\}$ be a database of images. Take $R(q) = (i_1, i_2, \dots, i_n)$ to be a permutation of the indices $1, \dots, n$ corresponding to the sorting of images retrieved for the query $q$. Let $\delta(x;q) = 1$ if $x$ is similar to $q$ and $\delta(x;q) = 0$ otherwise. The average precision of the first $k \leq n$ entries in the permutation $R(q)$ is
\[ AP_k(q) = \frac{\sum_{j=1}^k P(x_{i_1}, \dots, x_{i_j}; q) \delta(x_{i_j}; q) }{\sum_{j=1}^k \delta(x_{i_j}; q) } \]
where
\[P(x_{i_1}, \dots, x_{i_j}; q) = \frac{1}{j}\sum_{l=1}^j \delta(x_{i_l}; q), \]
is the precision up to $j$.
For a given set $Q$ of query points, the mean average precision at $k$ is then defined as
\[ \texttt{mAP@k} = \frac{1}{|Q|} \sum_{q \in Q} AP_k(q). \]
We compute \texttt{mAP@k} with $k=1000$ for ImageNet and $k=5000$ for the remaining datasets, as is common in the field \cite{hashnet2017}. To evaluate a hashing scheme, we take $R(q)$ to be the ordering given by the Hamming distance between the hashes of $q$ and the hashes of the images in $D$ with ties broken using the cosine distance of the embeddings. High values of \texttt{mAP@k} imply most items returned in the first positions are similar to the query; thus, higher \texttt{mAP@k} is better.

\vspace{1em}

\textbf{Deep hashing benchmarks.}
To evaluate the performance of \HQ\ quantization, we consider its effect on six state-of-the-art benchmarks from the field: DPSH \cite{dpsh2015}, DHN \cite{dhn2016}, HashNet \cite{hashnet2017}, DCH \cite{dch2018}, WGLHH \cite{weighted2021}, and \Hyp \cite{hyp22022}. The first three use similarity measures based on the inner product while the last three use cosine similarity. We also consider the classical Cosine Embedding Loss (CEL) \cite{cel1993}, which is a classical metric learning algorithm. Each of these methods define a different loss function following \eqref{eq:two_terms} (see \Cref{subsec:sm_dh_losses} for details). 

\vspace{1em}

\textbf{Quantization strategies.}
We compare how deep hashing benchmarks fare using the following strategies:
\begin{itemize}
    \item no quantization, where we take $\lambda = 0$ in \eqref{eq:two_terms};
    \item the original quantization penalty, which is obtained by taking $\lambda > 0$ in \eqref{eq:two_terms} (see details on  \Cref{subsec:sm_dh_losses});
    \item \HQ, as described in Algorithm \ref{alg:h2q};
    \item ITQ \cite{gong2012iterative}, where an orthogonal transformation is found through an iterative optimization process;
    \item HWSD \cite{doan2022one}, where the $L_Q$ term in \eqref{eq:two_terms} is replaced by their proposed quantization loss based on the sliced Wasserstein distance.
\end{itemize}

\vspace{1em}

\textbf{Neural network architectures.} Following the learning to hash literature, we employed both AlexNet \cite{alexnet} and VGG-16 \cite{vgg16}, and
adapted the architectures by replacing the last fully connected layers with softmax by a single fully connected layer with no activation. The weights of the hashing layer were initialized following a centered normal with deviation $0.01$ and the bias initialized with zeros. The weights and bias of all other layers were initialized using the pre-trained weights from \texttt{IMAGENET1K\_V1} available on \href{https://pytorch.org/vision/main/models.html}{torchvision}. Methods that use a quantization penalty (\ie, $\lambda>0$ in (\ref{eq:two_terms})) require a final $\tanh$ activation layer constraining the embeddings to $(0,1)^k$ to enforce quantization.

\vspace{1em}

\textbf{H²Q Optimization.} To solve (\ref{eq:orthogonal_transformation_optimization}), we optimize the objective function over $v_1, \ldots, v_k \in \R^{k}$ using (\ref{eq:householder_decomp}), which can be thought of as trainable vectors. We perform SGD using the Adam optimizer, employing the matrix multiplication algorithm in \cite{fasth} to quickly evaluate $U \bar{f}_{i}$ in each batch.

\vspace{1em}

\textbf{Hyperparameters.} Each benchmarks uses the recommended set of hyperparameters recommended by the respective authors (see \Cref{subsec:sm_dh_losses}; when it is not available, they are picked using a validation set). For all methods, we used the Adam optimizer \cite{kingma2014adam} with learning rate of $10^{-5}$ for all pre-trained layers and $10^{-4}$ for the hash layer. For every set of hyperparameters and quantization strategy, every method was run four times with different initializations; the final metric is the average of the four runs. We train the Householder transformation with the $L_2$ loss in \eqref{eq:orthogonal_transformation_optimization} and a learning rate of $0.1$ using 300 epochs and a batch size of 128 (for other choices, see \Cref{sec:ablation}).

\begin{table*}[ht]\centering
\small
\begin{tabular}{l|c@{\hskip .04in}c@{\hskip .04in}c@{\hskip .04in}c|c@{\hskip .04in}c@{\hskip .04in}c@{\hskip .04in}c|c@{\hskip .04in}c@{\hskip .04in}c@{\hskip .04in}c|c@{\hskip .04in}c@{\hskip .04in}c@{\hskip .04in}c}
& \multicolumn{4}{c|}{CIFAR 10} & \multicolumn{4}{c|}{NUS WIDE} & \multicolumn{4}{c|}{MS COCO} & \multicolumn{4}{c}{ImageNet} \\ \hline
number of bits ($k$) & 16 & 32 & 48 & 64 & 16 & 32 & 48 & 64 & 16 & 32 & 48 & 64 & 16 & 32 & 48 & 64\\ \hline
ADSH & 56.7 & 71.8 & 77.3 & 79.7 & 74.8 & 78.4 & 79.8 & 80.3 & 57.9 & 61.1 & 63.7 & 65.0 & 5.2 & 8.3 & 13.4 & 23.2\\
CEL & 79.8 & 81.0 & 81.7 & 81.3 & 79.4 & 80.3 & 80.7 & 80.7 & 64.4 & 66.3 & 67.5 & 68.4 & 51.8 & 52.5 & 53.7 & 45.6\\
DHN & 81.2 & 81.1 & 81.1 & 81.3 & 80.6 & 81.3 & 81.6 & 81.7 & 66.8 & 67.3 & 69.2 & 69.4 & 25.1 & 32.4 & 35.7 & 38.2\\
DCH & 80.2 & 80.1 & 80.0 & 79.8 & 78.4 & 79.1 & 79.1 & 79.8 & 63.8 & 66.2 & 67.1 & 66.7 & \textbf{58.2} & 58.8 & 58.9 & 60.4\\
DPSH & 81.2 & 81.2 & 81.5 & 81.1 & 81.0 & 81.9 & 82.1 & 82.1 & 68.0 & 71.2 & 71.6 & 72.4 & 36.5 & 42.2 & 46.0 & 49.9\\
HashNet & 80.8 & 82.1 & 82.3 & 82.3 & 79.8 & 81.5 & 82.2 & 82.7 & 62.9 & 67.3 & 68.2 & 70.2 & 41.2 & 54.3 & 58.8 & \textbf{62.5}\\
WGLHH & 79.6 & 80.0 & 80.2 & 79.4 & 79.9 & 80.7 & 80.1 & 80.5 & 66.3 & 67.0 & 67.7 & 67.2 & 55.3 & 57.1 & 57.0 & 56.8\\
HyP² & 80.5 & 81.1 & 81.7 & 81.8 & 81.9 & 82.5 & 83.1 & 83.0 & 71.9 & 74.1 & 74.8 & 74.9 & 54.1 & 56.9 & 57.7 & 56.5\\
HyP² + H²Q & \textbf{82.3} & \textbf{82.5} & \textbf{82.9} & \textbf{83.1} & \textbf{82.5} & \textbf{83.2} & \textbf{83.4} & \textbf{83.3} & \textbf{73.9} & \textbf{75.4} & \textbf{75.9} & \textbf{75.7} & 57.3 & \textbf{60.7} & \textbf{61.5} & 60.6\\
\end{tabular}

\caption{\texttt{mAP@k} of each benchmark over AlexNet, along with with H²Q improvement over HyP². Numbers in bold indicate that best metric overall for a given choice of method and number of bits. Note H²Q achieves the best performance in all cases, except for two, where it is the second best available (and always better than plain HyP², with improvements of up to 7.4\%).}
\label{tab:sota_CNNF_alexnet}
\end{table*}

\begin{table*}[ht]\centering
\small
\begin{tabular}{l|c@{\hskip .04in}c@{\hskip .04in}c@{\hskip .04in}c|c@{\hskip .04in}c@{\hskip .04in}c@{\hskip .04in}c|c@{\hskip .04in}c@{\hskip .04in}c@{\hskip .04in}c|c@{\hskip .04in}c@{\hskip .04in}c@{\hskip .04in}c}
& \multicolumn{4}{c|}{CIFAR 10} & \multicolumn{4}{c|}{NUS WIDE} & \multicolumn{4}{c|}{MS COCO} & \multicolumn{4}{c}{ImageNet} \\ \hline
number of bits ($k$) & 16 & 32 & 48 & 64 & 16 & 32 & 48 & 64 & 16 & 32 & 48 & 64 & 16 & 32 & 48 & 64\\ \hline
CEL ($\lambda=0$) & 79.8 & 81.0 & 81.7 & 81.3 & 79.4 & 80.3 & 80.7 & 80.7 & 64.4 & 66.3 & 67.5 & 68.4 & 51.8 & 52.5 & 53.7 & 45.6\\
CEL + H²Q & \textbf{82.2} & \textbf{82.4} & \textbf{82.7} & \textbf{82.4} & \textbf{80.6} & \textbf{81.9} & \textbf{82.2} & \textbf{82.3} & \textbf{66.4} & \textbf{68.5} & \textbf{69.6} & \textbf{70.2} & \textbf{54.6} & \textbf{55.0} & \textbf{56.1} & \textbf{48.4}\\[.4em]
DHN ($\lambda=0$) & 78.9 & 79.5 & 78.7 & 79.4 & 79.6 & 80.4 & 80.9 & 81.3 & 62.9 & 65.5 & 66.5 & 67.4 & 24.1 & 31.8 & 34.2 & 36.7\\
DHN + H²Q & \textbf{80.5} & \textbf{80.7} & \textbf{79.9} & \textbf{80.5} & \textbf{80.5} & \textbf{81.4} & \textbf{81.5} & \textbf{82.0} & \textbf{64.6} & \textbf{67.2} & \textbf{68.0} & \textbf{68.9} & \textbf{26.0} & \textbf{34.4} & \textbf{36.4} & \textbf{38.8}\\[.4em]
DCH ($\lambda=0$) & 78.3 & 77.5 & 77.3 & 76.3 & 78.8 & 78.9 & 78.5 & 78.6 & 62.8 & 64.1 & 64.2 & 64.3 & 50.9 & 49.6 & 48.5 & 46.5\\
DCH + H²Q & \textbf{81.6} & \textbf{80.3} & \textbf{80.1} & \textbf{79.4} & \textbf{79.6} & \textbf{79.9} & \textbf{79.7} & \textbf{80.1} & \textbf{64.3} & \textbf{66.0} & \textbf{66.1} & \textbf{66.1} & \textbf{55.0} & \textbf{53.5} & \textbf{51.9} & \textbf{50.0}\\[.4em]
WGLHH ($\lambda=0$) & 78.3 & 76.9 & 75.6 & 76.1 & 79.4 & 79.6 & 79.4 & 78.8 & 64.4 & 64.0 & 64.0 & 64.0 & 49.8 & 46.5 & 47.6 & 48.4\\
WGLHH + H²Q & \textbf{81.6} & \textbf{80.9} & \textbf{80.5} & \textbf{80.2} & \textbf{81.0} & \textbf{81.7} & \textbf{81.2} & \textbf{81.4} & \textbf{66.2} & \textbf{66.5} & \textbf{66.8} & \textbf{66.7} & \textbf{54.4} & \textbf{52.8} & \textbf{53.9} & \textbf{54.7}\\[.4em]
HyP² ($\lambda=0$) & 80.5 & 81.1 & 81.7 & 81.8 & 81.9 & 82.5 & 83.1 & 83.0 & 71.9 & 74.1 & 74.8 & 74.9 & 54.1 & 56.9 & 57.7 & 56.5\\
HyP² + H²Q & \textbf{82.3} & \textbf{82.5} & \textbf{82.9} & \textbf{83.1} & \textbf{82.5} & \textbf{83.2} & \textbf{83.4} & \textbf{83.3} & \textbf{73.9} & \textbf{75.4} & \textbf{75.9} & \textbf{75.7} & \textbf{57.3} & \textbf{60.7} & \textbf{61.5} & \textbf{60.6}\\[.4em]
\end{tabular}
\caption{\texttt{mAP@k} improvements over learning to hash benchmarks on AlexNet without quantization vs using H²Q. In all cases the performance metric increased. The overall average improvement is 3.6\%, and the maximum improvement is 13.5\%.}
\label{tab:improve_CNNF_alexnet}
\end{table*}

\textbf{Experiment Design.} For every dataset, every deep hashing method, every neural network architecture and bit size $k\in \{16, 32, 48, 64\}$, we repeat the following experiment:
\begin{enumerate}
    \item The deep hashing method is trained as proposed in its original publication, \ie, using quantization penalty ($\lambda > 0$) and the $\tanh$ activation layer, when applicable.
    \item \label{step2} A version of the deep hashing method is trained with no quantization penalty (\ie, $\lambda = 0$) and no $\tanh$ activation.
    \item The deep hashing method is also trained using the quantization term $L_Q$ given by the HWSD method.
    \item \HQ\ is trained on top of the embedding map learned by the deep hashing method considered in step \ref{step2} above. The same is done for ITQ.
\end{enumerate}
Each combination of deep hashing method, dataset, architecture and number of bit is executed four times and the average \texttt{mAP@k} is obtained and presented below.

\subsection{Improvements over Existing Benchmarks}
\label{sec:improve_sota}

Table \ref{tab:sota_CNNF_alexnet} shows that applying the \HQ\ quantization strategy to \Hyp\ \cite{hyp22022}, a state-of-the-art benchmark, is able to improve it in every case considered, and surpass all other deep hashing algorithms using the AlexNet architecture (see \Cref{sm-tab:sota_CNNF_vgg16_SM} for VGG-16). For all number of bits considered and all datasets, \HQ\ pushes \Hyp\ to have the best \texttt{mAP@k} metric in all but two cases, in which case it is second best. The improvement over \Hyp\ can often be significant, up to 7.4\%.

\begin{figure*}[ht]
    \centering
    \includegraphics[width=\linewidth]{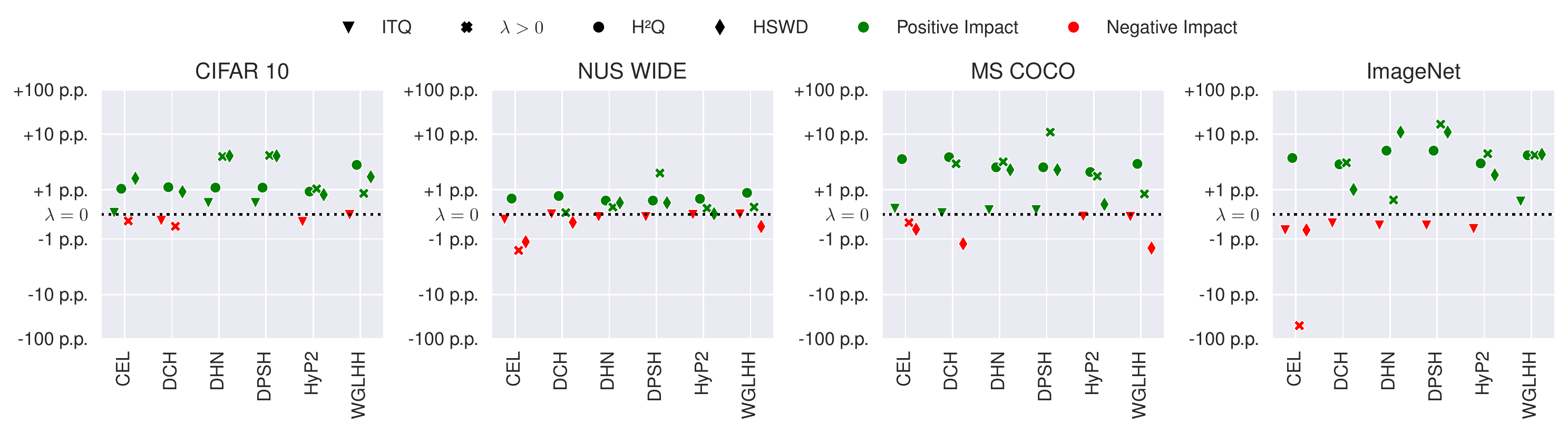}
    \caption{Variation of \texttt{mAP@k} in percentage points (p.p.) of each quantization strategy relative to no quantization ($\lambda = 0$) on VGG-16 with $k=16$ bits. Unlike ITQ, HWSD and quantization penalty terms ($\lambda > 0$), \HQ\ always increases the performance metric. }
    \label{fig:other_quantizations}
\end{figure*}

\subsection{Improvements over Similarity-based Losses}
\label{sec:uniform_improvements_on_similarity_losses}

An important consideration regarding a quantization strategy is whether it can negatively affect the performance of the learned embedding. Table \ref{tab:improve_CNNF_alexnet} shows that \HQ\ is always able to improve each of our deep hashing benchmarks relative to the same method without quantization when using the AlexNet architecture. The average improvement is 3.6\%, while the maximum improvement is 13.5\%. \Cref{sm-tab:improve_CNNF_vgg16_SM} contains the results for the VGG-16 architecture, with the same conclusion. (Note that DPSH has the same similarity term as DHN, thus we do not report it to avoid redundancy.)

\subsection{Comparison Against Other Quantization Benchmarks}
\label{sec:other_quantizations}

On the other hand, \Cref{fig:other_quantizations} shows that, for the VGG-16 network and $k=16$ bits, all the other quantization strategies considered sometimes hurt performance relative to running the same method with no penalization ($\lambda=0$). \Cref{sm-sec:comp_quant_strat} contains similar results for other architectures and number of bits. While \HQ\ systematically improves the metrics and is competitive with other methods, ITQ frequently reduces the \texttt{mAP@k}; when it provides improvements, they are usually smaller than those of \HQ. Using the original penalization terms ($\lambda>0$) is often more competitive than ITQ, but can also reduce the \texttt{mAP@k} and does not outperform \HQ\ consistently. HWSD performs better than ITQ, but can also reduce the \texttt{mAP@k} and does not outperform \HQ\ consistently.

\subsection{Ablation Study}
\label{sec:ablation}

\begin{figure*}
    \centering
    \includegraphics[width=\linewidth]{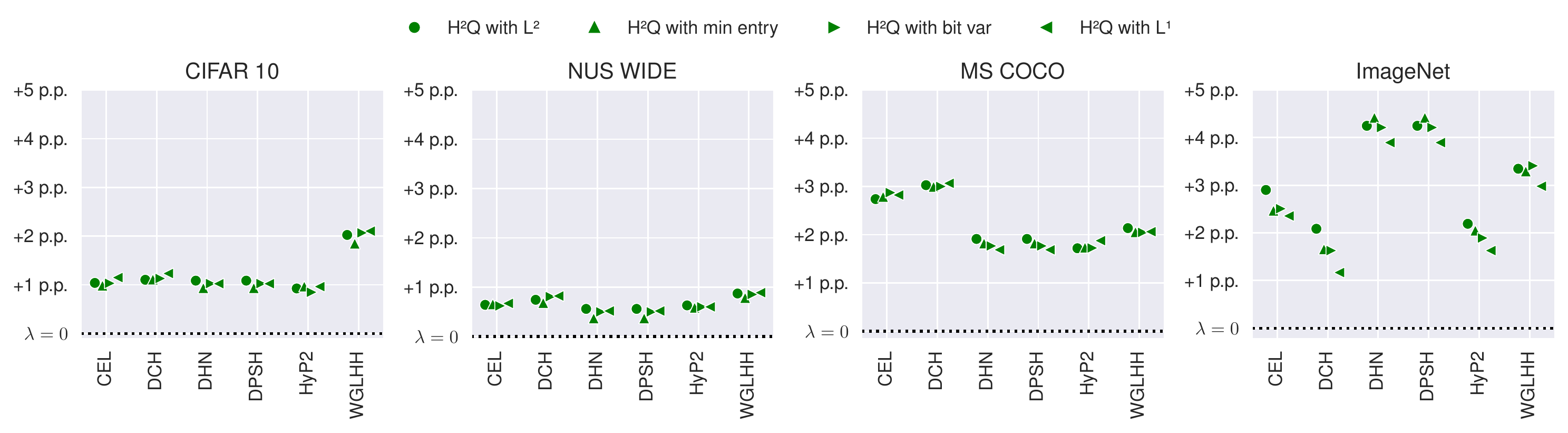}
    \caption{Increase in \texttt{mAP@k} in percentage points (p.p.) after using \HQ\ as a quantization strategy with different underlying losses, relative to no quantization ($\lambda=0$). On VGG-16 with $k=16$ bits. The increase is consistently positive and generally uniform across loss functions.} 
    \label{fig:ablation}
\end{figure*}

In \eqref{eq:orthogonal_transformation_optimization}, we proposed optimizing the $L_2$ distance between the rotated embedding $U f_\theta(x_i)$ and its quantization $\sign(U f_\theta(x_i))$. 
Still, since we are optimizing via SGD, it is easy to consider any other differentiable metric in the optimization. In this section, we study the effect of using three other reasonable loss functions beyond $L_2$ in \eqref{eq:orthogonal_transformation_optimization}. For simplicity, let $z_i = z_i(U) = U f_\theta(x_i)$ for every $i=1,\dots,n$ and $z_i^j$ be the j-th coordinate of $z_i$.

$\mathbf{L_1}:$ The alternative is to simply replace the $L_2$ distance by $L_1$. For that we change \eqref{eq:orthogonal_transformation_optimization} to
\begin{equation}
    \label{eq:L1_problem}
    \min_{v_1, \dots, v_k \in \mathbb{R}^k }\frac{1}{n}\sum_{i=1}^n \| z_i - \sign\left( z_i \right) \|_1.
\end{equation}
As before, this simply measures the distance between the rotated embedding to its quantization.

\textbf{min entry:} Another strategy is to ensure that no coordinate of $z_i$ is too close to $0$. Since the LogSumExp (LSE) function is a smooth version of the maximum we can solve
\begin{equation}
    \label{eq:min_entry}
    \min_{v_1, \dots, v_k \in \mathbb{R}^k }\frac{1}{n}\sum_{i=1}^n \mbox{LSE}(-(z_i)^2),   
\end{equation}
where the square in $(z_i)^2$ is done entry-wise.

\textbf{bit var:} Another alternative is to guarantee with high probability that no coordinate of $z_i$ is too small. Let $\xi$ be a random vector in $\R^k$ with i.i.d.\ entries drawn from a law with cumulative distribution function (CDF) $F$. We minimize the sum of the variance of the components of $\sign(z_i + \xi)$: 
\begin{equation}
    \label{eq:bit_var_loss}
    \min_{v_1, \dots, v_k \in \mathbb{R}^k }\frac{1}{n}\sum_{i=1}^n \sum_{j=1}^k F(z_i^j)(1 - F(z_i^j)).   
\end{equation}
The goal is to ensure that $z_i$ is far from the regions where there might be a change of sign.
In our experiments we take $F = (1+e^{-x})^{-1}$, the CDF of the logistic distribution.

The comparisons between the $L_2$ loss and the other three choices are shown in \Cref{fig:ablation}. 
While $L_2$ generally outperforms other alternatives, one might still prefer them for specific datasets and similarity losses. For example, WGLHH in CIFAR 10 attains better results using \eqref{eq:min_entry} or \eqref{eq:bit_var_loss}. The fact that there is little variation across loss function performance suggests that choosing the $L_2$ loss is close to the best possible alternative for orthogonal transformations.

\subsection{Computational Cost}
\label{sec:computational_cost}

\begin{figure}
    \centering
    \includegraphics[width=.6\linewidth]{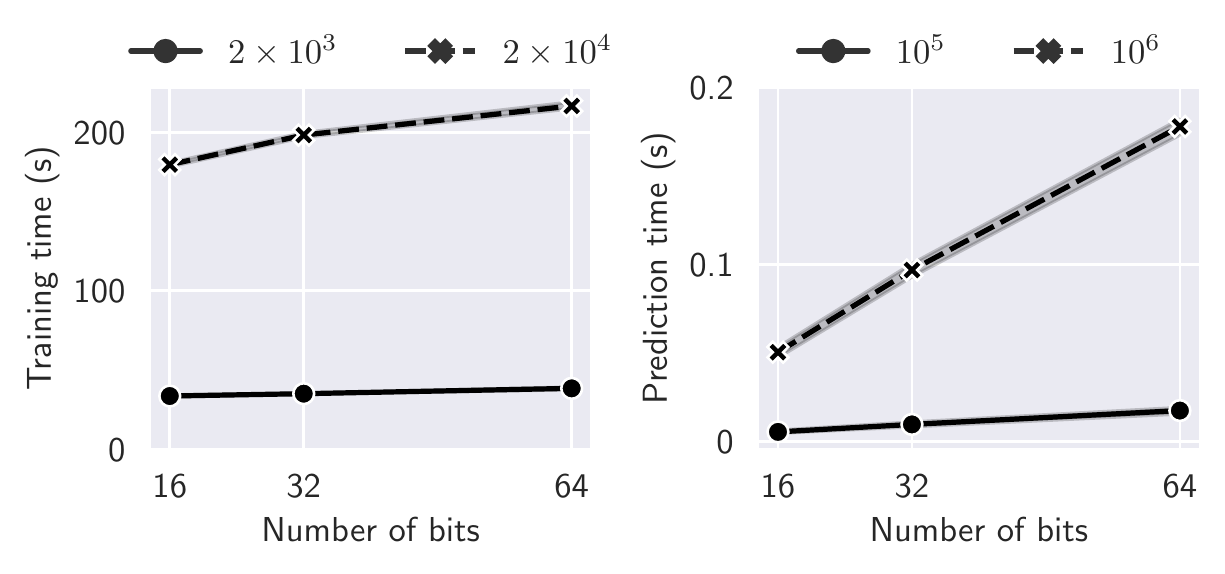}
    \caption{Training and prediction times for different number of bits. We vary the size of the training fold in $\{2\times 10^3, 2\times 10^4\}$ and predict $\{10^5, 10^6\}$ hashes, respectively. Training time is at most 3 minutes and prediction time is less than a second.}
    \label{fig:comp_time}
\end{figure}

A final concern regarding \HQ\ is the speed at which this quantization strategy can be executed. First, note that the only trainable parameters are the vectors $v_1, \cdots, v_k \in \mathbb{R}^k$ from \eqref{eq:householder_decomp}, a total of $k^2$ parameters. Using 32-bits float representation that gives 1kB, 4kB and 16kB for 16, 32 and 64 bits, respectively, and easily fitting in the L1 cache of any modern CPU. Moreover, a training set of 20,000 embeddings fits in 5MB in the 64-bits case, which fits in L2 cache. Thus, training can be performed without the need of a GPU. Indeed, it is faster to train on CPU to reduce latency. 

Figure \ref{fig:comp_time} shows the computational times for training and prediction, using the $L_2$ loss \eqref{eq:orthogonal_transformation_optimization}, a batch size of 128 and 300 epochs. Times were evaluated on a i9-13900K CPU (with L1 cache of 1.4MB, L2 of 34MB) with 128GB of RAM. Even on the 64 bits case, training on a CPU with 20,000 sample points takes around 3 minutes and predicting the hashes for $10^6$ sample points takes only 0.2 seconds. 

\section{Conclusion}

We propose \HQ, a novel quantization strategy for deep hashing that decomposes the learning to hash problem in two separate steps. First, perform similarity learning in the embedding space; second, optimize an orthogonal transformation to reduce quantization error. This stands in contrast to the prevalent quantization strategy, which jointly optimizes for both the embedding and the quantization error. Because \HQ\ relies on orthogonal transformations, it is able to minimize the quantization error while degrading the embedding by exploring an invariance in the similarity learning setup. \HQ\ is fast and hyperparameter-free, and uses a Householder matrix characterization of the orthogonal group along with SGD for quick computation. Unlike other popular quantization strategies, \HQ\ never hurts the embedding performance, and is able to significantly improve current benchmarks to deliver state-of-the-art performance in image retrieval tasks. A future avenue of work is exploring how \HQ\ can be combined with state-of-the-art metric learning methods on domains other than image retrieval. We hope our work encourages researchers to explore further invariances available in common loss functions to improve on learning to hash techniques and, more broadly, large-scale similarity search.

\section*{Reproducibility}

The companion code for this work can be found at \texttt{https://github.com/Lucas-Schwengber/h2q}.

\printbibliography

\newpage
\renewcommand{\thetable}{A\arabic{table}}
\appendix
\appendixhead

\section{Setup details}

We carefully reproduce the same parameters and training conditions proposed by the original paper of each similarity loss we used in our experiments. Meanwhile, we also needed to standardize our experimental setup. This section exhaustively describes the choices we made.

\subsection{Architectures}

We adapted two well known CNNs for the image classification problem to learning to hash. The goal of each CNN will be to learn a embedding map that maps an image into a point in $\R^k$. The hash is obtained taking the sign point-wise on the embedding. The CNNs adapted were the AlexNet and the VGG-16.

Since the original architectures are built to perform classification in ImageNet their last layer is a FC with output dimension $1000$. We replace this last layer with a new layer with output dimension $k$. We used the implementation of \cite{hashnet2017}, which is available in their GitHub \href{https://github.com/thuml/HashNet/tree/master}{repository}.

\subsection{Optimization}
\subsubsection{Learning the Embeddings}
We use the Adam optimizer with a learning rate of $10^{-5}$ both to train AlexNet and VGG-16. We also set a weight decay of $5\times 10^{-4}$ for all losses, except for WGLHH that uses $10^{-4}$. This is in line with the recommendation by the original papers of each loss.

The learning rate of the last layer of both adapted architectures (the hash layer), which is not pretrained, is set as $10^{-4}$. The proxies of the HyP² loss are trained with a learning rate of $10^{-3}$.

We also use the validation set to perform validation at the end of each epoch. During the validation we evaluated the $\texttt{mAP}$ metric choosing from the validation set in a set of $100$ queries and a set of $900$ possible retrieval images. To mitigate the effects of randomness we took the average $\texttt{mAP}$ of five random splits. We implemented an early stopping callback to halt the training if there is no observed improvement in the $\texttt{mAP}$ over a span of 20 epochs. The maximum number of epochs was set to 100.

\subsubsection{Learning the Orthogonal Transformations}

The orthogonal transformations are all trained using the Householder parametrization with the Adam optimizer, as discussed in the main text. The learning rate used for each of the four loss functions presented are different. The $L_2$, $L_1$ and the min entry losses all use $0.1$, but the bit var loss uses $0.01$. In all cases we use a batch size of $128$ and $300$ epochs.

\subsection{Deep Hashing Benchmark Losses}
\label{subsec:sm_dh_losses}

Let $n$ be the size of the mini-batch, $o_1, \dots, o_n \in \R^l$ be the embeddings of images $x_1, \dots, x_n$ and let $s_{ij} = 1$ if $x_i$ and $x_j$ are similar and $s_{ij} = 0$ otherwise. The cosine-similarity between $o_i$ and $o_j$ will be denoted by
\[ c_{ij} = \frac{\langle o_i, o_j \rangle}{\|o_i\|_2 \| o_j\|_2}, \]
and the generalized hamming distance by
\[ d_{ij} = \frac{k}{2}(1-c_{ij}). \]

Another important quantity is $w_{ij}$. This quantity aims to correct the unbalance between the amount of similar and dissimilar pairs. We define:
\[ w_{ij} = \frac{s_{ij}}{p} + \frac{1-s_{ij}}{1-p}. \]
where $p$ is the percentage of similar pairs in the training set.

Let $C$ be the number of classes of a dataset and $y_i \in \{0,1\}^C$ be the one hot labels of $x_i$. On CIFAR 10 and ImageNet each $x_i$ belong to one and only one class, while in NUS WIDE and MS COCO each $x_i$ can be in more than one class at the same time. In both cases the similarity between $i$ and $j$ is given by:
\[ s_{ij} = \textbf{1}_{ [\langle y_i, y_j \rangle > 0 ]}. \]

We now specialize the expression of the loss function of each benchmark following \eqref{eq:gen_sim_loss_func}.

\subsubsection{Cosine Embedding Loss (CEL) \cite{cel1993}} For this loss,
\[ L_S = \frac{1}{n^2} \sum_{i,j} s_{ij}\left(1- c_{ij} \right) + (1-s_{ij}) \left(c_{ij} - \delta \right)_+ \]
and
\[ L_Q = \frac{1}{n} \sum_{i=1}^n \| o_i - h_i \|_2^2. \]

The original CEL does not have a penalty term since it is used for metric learning. We add a penalty term to provide comparisons with such type of quantization strategy (\ie $\lambda > 0$).
The main parameter is the margin $\delta$ and is taken from \cite{deltavalue}, the batch size is 128. The penalty used was $\lambda = 0.01$.

\subsubsection{Deep Cauchy Hashing (DCH) \cite{dch2018}}
For this loss,
\[ L_S = \frac{1}{n^2} \sum_{i,j} w_{ij}\left( s_{ij}\log\frac{d_{ij}}{\gamma} + \log\frac{1+\gamma}{d_{ij}} \right)\]
and
\[ L_Q = \frac{1}{n} \sum_{i=1}^n \log\left( 1 + \gamma^{-1}\frac{k}{2}\left(1 - \frac{\langle o_i, h_i \rangle}{ \|o_i\|_2\|h_i\|_2 }\right) \right). \]

We let $\gamma = 10$ and a batch size of 256, as recommended in their paper (see Figure 5 of \cite{dch2018}). The penalty $\lambda$ was selected between the values $10^{-5}$, $10^{-3}$, and $10^{-1}$ using the validation set.

\subsubsection{Deep Hashing Network (DHN) \cite{dhn2016}}
For this loss,
\[ L_S = \frac{1}{n^2} \sum_{i,j} s_{ij}\langle o_i,o_j\rangle + \log\frac{1}{1+e^{-\langle o_i,o_j\rangle}}\]
and
\[ L_Q = \frac{1}{n} \sum_{i=1}^n \sum_{l=1}^k \log \frac{\cosh | o_{il} |}{e} . \]

We let a batch size of 64, as recommended in their paper. The penalty $\lambda$ was selected between the values $10^{-5}$, $10^{-3}$, and $10^{-1}$ using the validation set.

\subsubsection{Deep Pairwise-Supervised Hashing  (DPSH) \cite{dpsh2015}}
DPSH has the same similarity loss as DHN,
\[ L_S = \frac{1}{n^2} \sum_{i,j} s_{ij}\langle o_i,o_j\rangle + \log\frac{1}{1+e^{-\langle o_i,o_j\rangle}},\]
but the quantization loss is different:
\[ L_Q = \frac{1}{n} \sum_{i=1}^n \| o_i - h_i \|_2^2. \]
We let a batch size of 128. The penalty $\lambda$ was selected between the values $10^{-2}$, $10^{-1}$, and $1$ using the validation set.

\subsubsection{ Weighted Gaussian Loss Hamming Hashing (WGLHH) \cite{weighted2021}}
For this loss,
\begin{align*}
    L_S = & \frac{1}{n^2} \sum_{i,j} a_{ij} w_{ij} s_{ij} \log\frac{2s_{ij}}{s_{ij}+\exp(-\alpha d_{ij}^2)} + \frac{1}{n^2} \sum_{i,j} a_{ij} w_{ij} \exp(-\alpha d_{ij}^2) \log\frac{2 \exp(-\alpha d_{ij}^2) }{s_{ij}+\exp(-\alpha d_{ij}^2)}.
\end{align*}
with
\[ a_{ij} = \exp\left( \frac{s_{ij} - c_{ij}}{2} \right). \]

The quantization loss is
\[ L_Q = \frac{1}{n} \sum_{i=1}^n \| o_i - h_i \|_2^2. \]

We let $\alpha=0.1$, a batch size of 64 and $\lambda = 0.001$, as recommended in the original paper.

\subsubsection{HyP² \cite{hyp22022}}
For this loss we start defining a first term that is based on the dissimilar pairs:
\[ L_D = \frac{\sum_{i,j} (1-s_{ij})(c_{ij}-\delta)_+ }{\sum_{i,j} (1-s_{ij})}. \]

In this case a list of $C$ proxies are also trained. Each proxy stands for a class $1,\dots,C$ and tries to capture the ``average embedding'' of a class. Let $p_1, \dots, p_C \in \R^k$ be the proxies. Define
\begin{align*}
    L_P = & - \frac{\sum_{i=1}^n \sum_{l=1}^C y_{il} \frac{\langle o_i, p_l \rangle}{\| o_i \| \| p_l \|} }{\sum_{i=1}^n \sum_{l=1}^C y_{il}  } + \frac{\sum_{i=1}^n \sum_{l=1}^C (1-y_{il}) \left(\frac{\langle o_i, p_l \rangle}{\| o_i \| \| p_l \|} - \delta \right)_+ }{\sum_{i=1}^n \sum_{l=1}^C (1-y_{il})  }
\end{align*}
Finally,
\[ L_S = L_P + \beta L_D \]
and
\[ L_Q = \frac{1}{n} \sum_{i=1}^n \| o_i - h_i \|_2^2. \]

The quantization loss $L_Q$ is not originally proposed in their paper since their paper uses no quantization. We add the $L_Q$ loss in our experiments only for the case $\lambda > 0$. For benchmarks we consider their original formulation. The parameter $\delta$ is taken from \cite{deltavalue}. We followed the recommended batch size of 100. $\beta$ was selected among the values in $[0.5, 0.75, 1.0, 1.25]$, the choice was performed on the validation set to maximize \texttt{mAP@k}. Since the method does not use a quantization penalty term $L_Q$, we set $\lambda = 0.01$, as it is close to the default value for other methods.

\subsubsection{HashNet \cite{hashnet2017}}

HashNet takes another approach to quantization: instead of adding a penalty, they take $\beta \to \infty$ and evaluate their loss in $\tanh(\beta o_i)$, thus making the quantization error
\[ \| \tanh(\beta o_i) - h_i \|_2 \]
small. This quantization strategy is quite interesting, but is out of the scope of our work. We use HashNet as a hashing benchmark, but we do not experiment with it adding penalties or rotations. Our implementation directly follows the code available in their GitHub \href{https://github.com/thuml/HashNet/tree/master}{repository}, the main parameter is $\beta$, which we take as
\[ \left(1 + \frac{b}{200}\right)^\frac{1}{2},\]
where $b$ is the number of batches iterated until a moment (this number accumulates every epoch). The batch size used was $256$.

\subsubsection{ADSH \cite{adsh2018}}

ADSH learns the hash functions in two separate ways. For the points on the database, the hash codes are learned directly, while for out-of-sample extensions the hash codes come from binarizing the output of a neural network. The full optimization problem is formulated as:

\begin{equation}
    \begin{split}
        \min_{V, \Theta} \sum_{i=1}^m \sum_{j=1}^n \left[ \tanh\left(F\left(\textbf{x}_i, \Theta \right) \right)^{\intercal} \textbf{v}_j - c S_{ij} \right]^2, \\
        \text{s.t. } V \in \{ -1, +1 \}^{n \times k},
    \end{split}
\end{equation}

where $n$ is the size of the database and $m$ is the size of a random subset of images from the database. Their learning algorithm iterate between doing gradient descent updates on $\Theta$ using the random subset of the database and then updating the hash codes $V$ for the database points directly using a formula from linear algebra. Since they use the whole database for training, to make fair comparisons with the other methods, which are trained on the training set, we restrict the training step to the training set. The binary representation of each image is obtained using the out-of-sample extension provided by the neural network architecture. 

As for the choice of hyper parameters, as in the original paper, we take the maximum number of iterations to be $T_{\text{out}} = 50$, the strength of the penalization to be $\gamma = 200$, and the number of epochs per iteration to be $T_{\text{in}} = 3$. Differently from the original paper, we took the sample size to be $|\Omega| = 5000$ and the mini-batch size to be $64$ as this seemed to give better results in our setting. We set the starting learning rate to $10^{-5}$.

\subsection{Quantization Benchmarks}

In our work we study four main quantization strategies:
\begin{itemize}
    \item The baseline is to use no quantization strategy, here we train using the the benchmark losses described in \Cref{subsec:sm_dh_losses} with the quantization penalty set to zero;
    \item The second is to use the original quantization strategy proposed by each deep hashing method, that is to train using some penalty $\lambda > 0$;
    \item We also test our proposed quantization strategy, which consists of learning a rotation on top of the baseline;
    \item We also use the Iterative Quantization strategy \cite{gong2012iterative}, which also tries to learn a rotation on top of the baseline, but using an iterative and less-efficient process;
    \item Finally, we also add the HSWD quantization strategy proposed in \cite{hoe2021one} as a benchmark. This strategy simply replaces the original quantization loss term by another fixed quantization loss and train the embedding with the added penalty.
\end{itemize}

We now describe some implementation details of the previously listed quantization strategies.

\subsection{Baseline and Original Quantization Strategies}
For the baseline we use no $\tanh$ activation layer and set $\lambda = 0$. All other parameters are as described in \Cref{subsec:sm_dh_losses}. For the original quantization strategies we use the $\tanh$ activation layer. The penalty parameter $\lambda$ and all the other are as described in \Cref{subsec:sm_dh_losses}.

\subsubsection{ITQ}
Our implementation of ITQ follows this GitHub \href{https://github.com/twistedcubic/learn-to-hash/blob/master/itq.py}{repository}. The ITQ is originally design to obtain the embeddings using PCA decomposition. To adapt the method to be a quantization strategy for deep hashing methods we use the embeddings learned by the baseline as input. The main parameter is the number of iterations, we take it to be $50$. We also experiment if centralizing the embeddings (as is usually done when PCA is performed) would increase the performance, but it was not the case.

\subsubsection{HSWD}
This method simply exchange the original quantization term $L_Q$ of the deep hashing methods by another quantization term. We adapted the original implementation from their GitHub \href{https://github.com/khoadoan106/single_loss_quantization/blob/main/python/losses/distributional_quantization_losses.py}{repository}. The main parameter is the quantization penalty used, we follow the recommendation presented in their code and use $\lambda = 0.1$, we also do not use the $\tanh$ activation. All the other parameters follows the default parameter listed in \Cref{subsec:sm_dh_losses}.

\section{Experimental Results}\label{sec:sm-experimental_results}

\begin{table*}[ht]\centering
\small
\begin{tabular}{l|c@{\hskip .04in}c@{\hskip .04in}c@{\hskip .04in}c|c@{\hskip .04in}c@{\hskip .04in}c@{\hskip .04in}c|c@{\hskip .04in}c@{\hskip .04in}c@{\hskip .04in}c|c@{\hskip .04in}c@{\hskip .04in}c@{\hskip .04in}c}
& \multicolumn{4}{c|}{CIFAR 10} & \multicolumn{4}{c|}{NUS WIDE} & \multicolumn{4}{c|}{MS COCO} & \multicolumn{4}{c}{ImageNet} \\ \hline
n bits & 16 & 32 & 48 & 64 & 16 & 32 & 48 & 64 & 16 & 32 & 48 & 64 & 16 & 32 & 48 & 64\\ \hline
ADSH & 56.7 & 71.8 & 77.3 & 79.7 & 74.8 & 78.4 & 79.8 & 80.3 & 57.9 & 61.1 & 63.7 & 65.0 & 5.2 & 8.3 & 13.4 & 23.2\\
CEL & 79.8 & 81.0 & 81.7 & 81.3 & 79.4 & 80.3 & 80.7 & 80.7 & 64.4 & 66.3 & 67.5 & 68.4 & 51.8 & 52.5 & 53.7 & 45.6\\
DHN & 81.2 & 81.1 & 81.1 & 81.3 & 80.6 & 81.3 & 81.6 & 81.7 & 66.8 & 67.3 & 69.2 & 69.4 & 25.1 & 32.4 & 35.7 & 38.2\\
DCH & 80.2 & 80.1 & 80.0 & 79.8 & 78.4 & 79.1 & 79.1 & 79.8 & 63.8 & 66.2 & 67.1 & 66.7 & \textbf{58.2} & 58.8 & 58.9 & 60.4\\
DPSH & 81.2 & 81.2 & 81.5 & 81.1 & 81.0 & 81.9 & 82.1 & 82.1 & 68.0 & 71.2 & 71.6 & 72.4 & 36.5 & 42.2 & 46.0 & 49.9\\
HashNet & 80.8 & 82.1 & 82.3 & 82.3 & 79.8 & 81.5 & 82.2 & 82.7 & 62.9 & 67.3 & 68.2 & 70.2 & 41.2 & 54.3 & 58.8 & \textbf{62.5}\\
WGLHH & 79.6 & 80.0 & 80.2 & 79.4 & 79.9 & 80.7 & 80.1 & 80.5 & 66.3 & 67.0 & 67.7 & 67.2 & 55.3 & 57.1 & 57.0 & 56.8\\
HyP² & 80.5 & 81.1 & 81.7 & 81.8 & 81.9 & 82.5 & 83.1 & 83.0 & 71.9 & 74.1 & 74.8 & 74.9 & 54.1 & 56.9 & 57.7 & 56.5\\
HyP² + ITQ & 80.2 & 80.7 & 81.4 & 81.8 & 81.6 & 82.5 & 83.0 & 83.0 & 71.6 & 73.9 & 74.8 & 74.8 & 53.5 & 56.7 & 57.6 & 56.3\\
HyP² + HSWD & 82.1 & 82.3 & 82.4 & 82.0 & 82.0 & 82.6 & 83.2 & 83.1 & 72.1 & 74.7 & 74.9 & 74.9 & 56.8 & 60.2 & 60.8 & 58.8\\
HyP² + H²Q & \textbf{82.3} & \textbf{82.5} & \textbf{82.9} & \textbf{83.1} & \textbf{82.5} & \textbf{83.2} & \textbf{83.4} & \textbf{83.3} & \textbf{73.9} & \textbf{75.4} & \textbf{75.9} & \textbf{75.7} & 57.3 & \textbf{60.7} & \textbf{61.5} & 60.6\\
\end{tabular}
\caption{\texttt{mAP@k} with AlexNet. We compare the best performance metrics of each benchmark with HyP² combined with H²Q.}
\label{sm-tab:sota_CNNF_alexnet_SM}
\end{table*}

\begin{table*}[ht]\centering
\small
\begin{tabular}{l|c@{\hskip .04in}c@{\hskip .04in}c@{\hskip .04in}c|c@{\hskip .04in}c@{\hskip .04in}c@{\hskip .04in}c|c@{\hskip .04in}c@{\hskip .04in}c@{\hskip .04in}c|c@{\hskip .04in}c@{\hskip .04in}c@{\hskip .04in}c}
& \multicolumn{4}{c|}{CIFAR 10} & \multicolumn{4}{c|}{NUS WIDE} & \multicolumn{4}{c|}{MS COCO} & \multicolumn{4}{c}{ImageNet} \\ \hline
n bits & 16 & 32 & 48 & 64 & 16 & 32 & 48 & 64 & 16 & 32 & 48 & 64 & 16 & 32 & 48 & 64\\ \hline
ADSH & 64.4 & 80.3 & 84.1 & 85.0 & 79.1 & 82.3 & 83.2 & 83.5 & 62.2 & 65.6 & 67.5 & 69.2 & 10.8 & 22.7 & 39.7 & 56.2\\
CEL & 85.1 & 85.6 & 85.8 & 85.7 & 82.7 & 83.1 & 83.3 & 82.8 & 73.3 & 75.7 & 76.2 & 76.4 & 69.9 & 72.8 & 74.4 & 72.5\\
DCH & 84.0 & 82.9 & 82.8 & 82.9 & 81.9 & 81.8 & 81.1 & 81.2 & 75.7 & 76.0 & 76.7 & 76.0 & \textbf{79.8} & 80.2 & 79.8 & 80.1\\
DHN & 86.2 & 86.6 & \textbf{86.9} & 86.6 & 83.2 & 83.6 & 83.6 & 83.7 & 68.6 & 70.7 & 69.9 & 71.5 & 45.4 & 54.4 & 59.1 & 62.3\\
DPSH & \textbf{86.4} & \textbf{86.8} & 86.5 & \textbf{86.7} & 84.5 & 85.0 & 85.2 & 85.5 & 77.4 & 79.0 & 79.7 & 79.1 & 61.6 & 68.8 & 71.6 & 74.0\\
HashNet & 85.7 & 86.1 & 86.4 & 86.2 & 83.2 & 84.1 & 84.6 & 85.0 & 68.3 & 72.2 & 74.3 & 75.5 & 65.0 & 74.7 & 79.2 & 80.8\\
WGLHH & 84.5 & 83.6 & 84.2 & 83.3 & 83.8 & 83.6 & 83.4 & 83.0 & 78.1 & 75.9 & 76.9 & 76.6 & 79.6 & 79.4 & 79.0 & 78.2\\
HyP² & 85.0 & 85.2 & 85.6 & 85.7 & 85.1 & 85.5 & 85.9 & 85.9 & 79.7 & 82.0 & 82.6 & 82.2 & 75.7 & 77.5 & 79.0 & 78.8\\
HyP² + ITQ & 84.7 & 85.0 & 85.6 & 85.5 & 85.1 & 85.6 & 85.8 & 85.9 & 79.6 & 81.9 & 82.6 & 82.1 & 75.1 & 77.0 & 78.4 & 78.8\\
HyP² + HSWD & 85.8 & 85.6 & 85.8 & 86.0 & 85.1 & 85.4 & 85.7 & 85.7 & 80.1 & 81.8 & 82.2 & 81.6 & 77.3 & 79.3 & 80.0 & 78.8\\
HyP² + H²Q & 86.0 & 86.3 & 86.4 & 86.4 & \textbf{85.7} & \textbf{86.0} & \textbf{86.1} & \textbf{86.1} & \textbf{81.4} & \textbf{82.7} & \textbf{83.1} & \textbf{82.4} & 77.9 & \textbf{80.6} & \textbf{81.5} & \textbf{81.5}\\
\end{tabular}
\caption{\texttt{mAP@k} with VGG-16. We compare the best performance metrics of each benchmark with HyP² combined with H²Q.}
\label{sm-tab:sota_CNNF_vgg16_SM}
\end{table*}

\begin{table*}[ht]\centering
\small
\begin{tabular}{l|c@{\hskip .04in}c@{\hskip .04in}c@{\hskip .04in}c|c@{\hskip .04in}c@{\hskip .04in}c@{\hskip .04in}c|c@{\hskip .04in}c@{\hskip .04in}c@{\hskip .04in}c|c@{\hskip .04in}c@{\hskip .04in}c@{\hskip .04in}c}
& \multicolumn{4}{c|}{CIFAR 10} & \multicolumn{4}{c|}{NUS WIDE} & \multicolumn{4}{c|}{MS COCO} & \multicolumn{4}{c}{ImageNet} \\ \hline
n bits & 16 & 32 & 48 & 64 & 16 & 32 & 48 & 64 & 16 & 32 & 48 & 64 & 16 & 32 & 48 & 64\\ \hline
CEL ($\lambda=0$) & 85.1 & 85.6 & 85.8 & 85.7 & 82.7 & 83.1 & 83.3 & 82.8 & 73.3 & 75.7 & 76.2 & 76.4 & 69.9 & 72.8 & 74.4 & 72.5\\
CEL + H²Q & \textbf{86.1} & \textbf{86.5} & \textbf{86.8} & \textbf{86.6} & \textbf{83.4} & \textbf{84.0} & \textbf{84.3} & \textbf{84.1} & \textbf{76.1} & \textbf{78.0} & \textbf{78.5} & \textbf{78.5} & \textbf{72.8} & \textbf{75.8} & \textbf{77.3} & \textbf{75.8}\\[.4em]
DHN ($\lambda=0$) & 83.1 & 85.5 & 85.9 & 85.8 & 82.9 & 83.2 & 83.6 & 83.3 & 66.3 & 69.5 & 70.0 & 71.4 & 44.8 & 55.6 & 59.9 & 62.5\\
DHN + H²Q & \textbf{84.2} & \textbf{86.2} & \textbf{86.8} & \textbf{86.8} & \textbf{83.4} & \textbf{83.9} & \textbf{84.1} & \textbf{83.9} & \textbf{68.2} & \textbf{71.7} & \textbf{72.7} & \textbf{73.7} & \textbf{49.1} & \textbf{59.9} & \textbf{64.1} & \textbf{66.1}\\[.4em]
DCH ($\lambda=0$) & 84.5 & 82.6 & 80.1 & 80.2 & 81.9 & 81.3 & 80.5 & 80.0 & 73.5 & 69.4 & 70.0 & 68.4 & 77.5 & 75.6 & 72.2 & 69.6\\
DCH + H²Q & \textbf{85.6} & \textbf{83.9} & \textbf{81.9} & \textbf{81.8} & \textbf{82.6} & \textbf{82.3} & \textbf{81.5} & \textbf{80.9} & \textbf{76.5} & \textbf{72.6} & \textbf{72.9} & \textbf{71.3} & \textbf{79.6} & \textbf{78.0} & \textbf{74.7} & \textbf{72.4}\\[.4em]
WGLHH ($\lambda=0$) & 83.6 & 81.9 & 80.9 & 80.6 & 83.5 & 82.9 & 82.6 & 82.4 & 77.2 & 73.8 & 74.9 & 75.5 & 76.2 & 75.1 & 75.4 & 77.0\\
WGLHH + H²Q & \textbf{85.6} & \textbf{84.3} & \textbf{83.1} & \textbf{83.3} & \textbf{84.4} & \textbf{83.9} & \textbf{83.4} & \textbf{83.4} & \textbf{79.4} & \textbf{76.0} & \textbf{77.1} & \textbf{77.6} & \textbf{79.6} & \textbf{78.5} & \textbf{78.4} & \textbf{79.2}\\[.4em]
HyP² ($\lambda=0$) & 85.0 & 85.2 & 85.6 & 85.7 & 85.1 & 85.5 & 85.9 & 85.9 & 79.7 & 82.0 & 82.6 & 82.2 & 75.7 & 77.5 & 79.0 & 78.8\\
HyP² + H²Q & \textbf{86.0} & \textbf{86.3} & \textbf{86.4} & \textbf{86.4} & \textbf{85.7} & \textbf{86.0} & \textbf{86.1} & \textbf{86.1} & \textbf{81.4} & \textbf{82.7} & \textbf{83.1} & \textbf{82.4} & \textbf{77.9} & \textbf{80.6} & \textbf{81.5} & \textbf{81.5}\\[.4em]
\end{tabular}
\caption{\texttt{mAP@k} with VGG-16. We show the effect of using H²Q on each method after training with no $tanh$ activation and no quantization term ($\lambda = 0$).}
\label{sm-tab:improve_CNNF_vgg16_SM}
\end{table*}

\subsection{Improvements over Existing Benchmarks}

Table \ref{sm-tab:sota_CNNF_alexnet_SM} is an extended version of Table \ref{tab:sota_CNNF_alexnet} of the main text, and Table \ref{sm-tab:sota_CNNF_vgg16_SM} is its analogous to the VGG-16 architecture. The \Hyp\  with our quantization strategy is the winning strategy in the majority of the cases also for the VGG-16.

\subsection{Improvements over Similarity-based Losses}

Here we extend the discussion provided in Section \ref{sec:uniform_improvements_on_similarity_losses} of the main text. Table \ref{sm-tab:improve_CNNF_vgg16_SM} is the analogous of Table \ref{tab:improve_CNNF_alexnet} of the main text, but for the VGG-16 architecture. As observed for the AlexNet, we uniformly improved the performance of similarity-based losses also for the VGG-16 architecture. We have an average improvement of $2.6\%$ and a maximum of $9.5\%$.



\begin{table*}[]\centering
\vspace*{-1cm}
\setlength\extrarowheight{-2pt}
\begin{tabular}{l@{\hskip .01in}|c@{\hskip .01in}c@{\hskip .01in}c@{\hskip .01in}c|c@{\hskip .01in}c@{\hskip .01in}c@{\hskip .01in}c|c@{\hskip .01in}c@{\hskip .01in}c@{\hskip .01in}c|c@{\hskip .01in}c@{\hskip .01in}c@{\hskip .01in}c}
& \multicolumn{4}{c|}{CIFAR 10} & \multicolumn{4}{c|}{NUS WIDE} & \multicolumn{4}{c|}{MS COCO} & \multicolumn{4}{c}{ImageNet} \\ \hline
n bits & 16 & 32 & 48 & 64 & 16 & 32 & 48 & 64 & 16 & 32 & 48 & 64 & 16 & 32 & 48 & 64\\ \hline
{\tiny CEL($\lambda=0$)} & {\tiny79.8$^{0.6}$} & {\tiny81.0$^{0.4}$} & {\tiny81.7$^{0.3}$} & {\tiny81.3$^{0.2}$} & {\tiny79.4$^{0.2}$} & {\tiny80.3$^{0.2}$} & {\tiny80.7$^{0.2}$} & {\tiny80.7$^{0.3}$} & {\tiny64.4$^{0.5}$} & {\tiny66.3$^{0.6}$} & {\tiny67.5$^{0.3}$} & {\tiny68.4$^{0.4}$} & {\tiny51.8$^{1.1}$} & {\tiny52.5$^{0.3}$} & {\tiny53.7$^{1.1}$} & {\tiny45.6$^{0.5}$}\\
{\tiny CEL+ITQ} & {\tiny79.8$^{0.5}$} & {\color{red}{\tiny80.6$^{0.4}$}} & {\color{red}{\tiny81.5$^{0.1}$}} & {\tiny81.3$^{0.3}$} & {\tiny79.4$^{0.1}$} & {\tiny80.5$^{0.3}$} & {\color{red}{\tiny80.4$^{0.2}$}} & {\tiny80.7$^{0.4}$} & {\tiny64.4$^{0.3}$} & {\tiny66.4$^{0.4}$} & {\tiny67.8$^{0.3}$} & {\tiny68.7$^{0.2}$} & {\color{red}{\tiny51.7$^{0.9}$}} & {\tiny53.0$^{0.5}$} & {\tiny53.7$^{0.8}$} & {\tiny45.7$^{0.5}$}\\
{\tiny CEL+$\lambda$} & {\tiny80.2$^{0.3}$} & {\color{red}{\tiny79.1$^{1.0}$}} & {\color{red}{\tiny77.8$^{0.4}$}} & {\color{red}{\tiny70.7$^{2.7}$}} & {\color{red}{\tiny79.2$^{0.2}$}} & {\color{red}{\tiny79.8$^{0.3}$}} & {\color{red}{\tiny79.6$^{0.1}$}} & {\color{red}{\tiny79.0$^{0.4}$}} & {\tiny66.1$^{0.3}$} & {\tiny67.1$^{0.2}$} & {\color{red}{\tiny67.3$^{0.5}$}} & {\color{red}{\tiny66.9$^{0.2}$}} & {\color{red}{\tiny19.1$^{0.8}$}} & {\color{red}{\tiny22.2$^{0.9}$}} & {\color{red}{\tiny23.3$^{0.1}$}} & {\color{red}{\tiny25.5$^{0.5}$}}\\
{\tiny CEL+HSWD} & {\tiny82.0$^{0.6}$} & {\tiny82.4$^{0.6}$} & \textbf{{\tiny82.8$^{0.5}$}} & \textbf{{\tiny82.9$^{0.6}$}} & {\tiny79.9$^{0.3}$} & {\tiny80.6$^{0.5}$} & {\tiny81.0$^{0.4}$} & {\tiny81.1$^{0.4}$} & {\color{red}{\tiny64.3$^{0.6}$}} & {\tiny66.3$^{0.6}$} & {\tiny67.6$^{0.8}$} & {\color{red}{\tiny68.3$^{0.5}$}} & {\color{red}{\tiny43.8$^{0.1}$}} & {\color{red}{\tiny48.1$^{0.3}$}} & {\color{red}{\tiny49.5$^{0.2}$}} & {\tiny46.9$^{0.5}$}\\
{\tiny CEL+H²Q($L_2$)} & {\tiny82.2$^{0.5}$} & {\tiny82.4$^{0.1}$} & {\tiny82.7$^{0.1}$} & {\tiny82.4$^{0.2}$} & {\tiny80.6$^{0.2}$} & \textbf{{\tiny81.9$^{0.2}$}} & {\tiny82.2$^{0.1}$} & {\tiny82.3$^{0.3}$} & {\tiny66.4$^{0.4}$} & {\tiny68.5$^{0.3}$} & {\tiny69.6$^{0.2}$} & {\tiny70.2$^{0.2}$} & \textbf{{\tiny54.6$^{0.7}$}} & \textbf{{\tiny55.0$^{0.4}$}} & \textbf{{\tiny56.1$^{0.7}$}} & {\tiny48.4$^{0.7}$}\\
{\tiny CEL+H²Q($L_1$)} & {\tiny82.3$^{0.6}$} & \textbf{{\tiny82.5$^{0.2}$}} & {\tiny82.7$^{0.2}$} & {\tiny82.5$^{0.3}$} & {\tiny80.6$^{0.2}$} & {\tiny81.8$^{0.2}$} & {\tiny82.1$^{0.1}$} & {\tiny82.3$^{0.3}$} & \textbf{{\tiny66.5$^{0.4}$}} & {\tiny68.4$^{0.3}$} & {\tiny69.6$^{0.2}$} & {\tiny70.2$^{0.2}$} & {\tiny54.3$^{0.7}$} & {\tiny54.8$^{0.5}$} & {\tiny55.9$^{0.6}$} & \textbf{{\tiny48.6$^{0.7}$}}\\
{\tiny CEL+H²Q(min)} & \textbf{{\tiny82.4$^{0.4}$}} & {\tiny82.4$^{0.2}$} & {\tiny82.7$^{0.2}$} & {\tiny82.5$^{0.4}$} & \textbf{{\tiny80.7$^{0.2}$}} & {\tiny81.8$^{0.3}$} & {\tiny82.2$^{0.1}$} & {\tiny82.3$^{0.3}$} & {\tiny66.4$^{0.3}$} & \textbf{{\tiny68.5$^{0.4}$}} & \textbf{{\tiny69.7$^{0.2}$}} & \textbf{{\tiny70.4$^{0.2}$}} & {\tiny54.2$^{0.7}$} & {\tiny54.6$^{0.7}$} & {\tiny56.0$^{0.7}$} & {\tiny48.5$^{0.5}$}\\
{\tiny CEL+H²Q(bit)} & {\tiny82.3$^{0.4}$} & {\tiny82.3$^{0.2}$} & {\tiny82.7$^{0.3}$} & {\tiny82.5$^{0.3}$} & {\tiny80.7$^{0.2}$} & {\tiny81.8$^{0.2}$} & \textbf{{\tiny82.2$^{0.1}$}} & \textbf{{\tiny82.3$^{0.3}$}} & {\tiny66.4$^{0.3}$} & {\tiny68.5$^{0.3}$} & {\tiny69.6$^{0.2}$} & {\tiny70.3$^{0.2}$} & {\tiny53.7$^{0.8}$} & {\tiny54.6$^{0.6}$} & {\tiny55.6$^{0.7}$} & {\tiny48.3$^{0.7}$}\\[.2em]
{\tiny DCH($\lambda=0$)} & {\tiny78.3$^{0.5}$} & {\tiny77.5$^{1.1}$} & {\tiny77.3$^{0.4}$} & {\tiny76.3$^{1.1}$} & {\tiny78.8$^{0.5}$} & {\tiny78.9$^{0.4}$} & {\tiny78.5$^{0.2}$} & {\tiny78.6$^{0.2}$} & {\tiny62.8$^{1.1}$} & {\tiny64.1$^{0.4}$} & {\tiny64.2$^{0.2}$} & {\tiny64.3$^{0.2}$} & {\tiny50.9$^{2.1}$} & {\tiny49.6$^{1.2}$} & {\tiny48.5$^{1.2}$} & {\tiny46.5$^{1.1}$}\\
{\tiny DCH+ITQ} & {\tiny78.4$^{0.5}$} & {\color{red}{\tiny77.1$^{0.9}$}} & {\color{red}{\tiny77.2$^{0.4}$}} & {\tiny76.5$^{1.0}$} & {\color{red}{\tiny78.5$^{0.4}$}} & {\color{red}{\tiny78.8$^{0.4}$}} & {\color{red}{\tiny78.2$^{0.3}$}} & {\tiny78.7$^{0.3}$} & {\color{red}{\tiny62.5$^{0.8}$}} & {\color{red}{\tiny63.8$^{0.4}$}} & {\tiny64.5$^{0.2}$} & {\tiny64.6$^{0.2}$} & {\tiny51.2$^{1.5}$} & {\color{red}{\tiny49.6$^{0.9}$}} & {\color{red}{\tiny48.2$^{1.3}$}} & {\tiny46.9$^{1.5}$}\\
{\tiny DCH+$\lambda$} & {\tiny80.2$^{0.5}$} & {\tiny80.1$^{0.4}$} & {\tiny80.0$^{0.4}$} & \textbf{{\tiny79.8$^{0.4}$}} & {\color{red}{\tiny78.4$^{0.2}$}} & {\tiny79.1$^{0.2}$} & {\tiny79.1$^{0.2}$} & {\tiny79.8$^{0.3}$} & {\tiny63.8$^{1.1}$} & \textbf{{\tiny66.2$^{0.6}$}} & \textbf{{\tiny67.1$^{1.6}$}} & \textbf{{\tiny66.7$^{0.7}$}} & \textbf{{\tiny58.2$^{0.9}$}} & \textbf{{\tiny58.8$^{0.7}$}} & \textbf{{\tiny58.9$^{1.5}$}} & \textbf{{\tiny60.4$^{0.7}$}}\\
{\tiny DCH+HSWD} & {\tiny81.0$^{0.9}$} & {\tiny80.0$^{0.5}$} & {\tiny79.3$^{0.1}$} & {\tiny78.1$^{2.3}$} & {\tiny79.2$^{0.4}$} & {\color{red}{\tiny78.9$^{0.2}$}} & {\color{red}{\tiny78.3$^{0.1}$}} & {\color{red}{\tiny78.2$^{0.2}$}} & {\tiny62.8$^{0.7}$} & {\color{red}{\tiny63.7$^{1.0}$}} & {\tiny64.3$^{0.5}$} & {\color{red}{\tiny63.9$^{0.6}$}} & {\tiny56.5$^{0.9}$} & {\tiny52.0$^{1.3}$} & {\tiny50.5$^{1.7}$} & {\tiny48.3$^{1.2}$}\\
{\tiny DCH+H²Q($L_2$)} & {\tiny81.6$^{0.5}$} & {\tiny80.3$^{0.6}$} & \textbf{{\tiny80.1$^{0.3}$}} & {\tiny79.4$^{0.6}$} & {\tiny79.6$^{0.4}$} & {\tiny79.9$^{0.2}$} & \textbf{{\tiny79.7$^{0.1}$}} & {\tiny80.1$^{0.3}$} & {\tiny64.3$^{1.3}$} & {\tiny66.0$^{0.2}$} & {\tiny66.1$^{0.1}$} & {\tiny66.1$^{0.1}$} & {\tiny55.0$^{1.3}$} & {\tiny53.5$^{1.0}$} & {\tiny51.9$^{1.1}$} & {\tiny50.0$^{0.7}$}\\
{\tiny DCH+H²Q($L_1$)} & {\tiny81.6$^{0.1}$} & {\tiny80.3$^{0.6}$} & {\tiny80.1$^{0.3}$} & {\tiny79.2$^{0.8}$} & {\tiny79.7$^{0.4}$} & {\tiny79.9$^{0.2}$} & {\tiny79.6$^{0.2}$} & {\tiny80.0$^{0.3}$} & {\tiny64.5$^{1.1}$} & {\tiny66.0$^{0.2}$} & {\tiny66.1$^{0.1}$} & {\tiny66.2$^{0.2}$} & {\tiny55.0$^{1.5}$} & {\tiny53.0$^{1.1}$} & {\tiny51.3$^{0.9}$} & {\tiny49.2$^{1.0}$}\\
{\tiny DCH+H²Q(min)} & {\tiny81.7$^{0.3}$} & \textbf{{\tiny80.4$^{0.6}$}} & {\tiny80.0$^{0.1}$} & {\tiny79.1$^{0.5}$} & \textbf{{\tiny79.7$^{0.3}$}} & \textbf{{\tiny80.0$^{0.2}$}} & {\tiny79.7$^{0.1}$} & {\tiny80.1$^{0.3}$} & \textbf{{\tiny64.7$^{1.2}$}} & {\tiny66.0$^{0.1}$} & {\tiny66.1$^{0.1}$} & {\tiny66.2$^{0.2}$} & {\tiny54.8$^{1.3}$} & {\tiny52.6$^{0.9}$} & {\tiny51.0$^{1.2}$} & {\tiny48.5$^{0.6}$}\\
{\tiny DCH+H²Q(bit)} & \textbf{{\tiny81.7$^{0.3}$}} & {\tiny80.2$^{0.4}$} & {\tiny79.6$^{0.3}$} & {\tiny78.9$^{0.6}$} & {\tiny79.7$^{0.4}$} & {\tiny80.0$^{0.2}$} & {\tiny79.7$^{0.1}$} & \textbf{{\tiny80.2$^{0.4}$}} & {\tiny64.6$^{1.1}$} & {\tiny65.9$^{0.2}$} & {\tiny66.1$^{0.1}$} & {\tiny66.1$^{0.2}$} & {\tiny54.7$^{1.3}$} & {\tiny52.6$^{1.0}$} & {\tiny50.3$^{0.9}$} & {\tiny48.0$^{0.9}$}\\[.2em]
{\tiny DHN($\lambda=0$)} & {\tiny78.9$^{1.1}$} & {\tiny79.5$^{0.5}$} & {\tiny78.7$^{0.6}$} & {\tiny79.4$^{0.4}$} & {\tiny79.6$^{0.2}$} & {\tiny80.4$^{0.3}$} & {\tiny80.9$^{0.3}$} & {\tiny81.3$^{0.2}$} & {\tiny62.9$^{1.5}$} & {\tiny65.5$^{1.4}$} & {\tiny66.5$^{3.1}$} & {\tiny67.4$^{2.5}$} & {\tiny24.1$^{0.6}$} & {\tiny31.8$^{0.5}$} & {\tiny34.2$^{0.7}$} & {\tiny36.7$^{1.3}$}\\
{\tiny DHN+ITQ} & {\tiny79.1$^{1.0}$} & {\color{red}{\tiny79.2$^{0.6}$}} & {\tiny78.7$^{1.0}$} & {\color{red}{\tiny79.3$^{0.4}$}} & {\color{red}{\tiny79.6$^{0.1}$}} & {\color{red}{\tiny80.4$^{0.1}$}} & {\tiny81.0$^{0.2}$} & {\tiny81.5$^{0.2}$} & {\tiny63.1$^{1.1}$} & {\tiny65.5$^{1.5}$} & {\tiny66.6$^{3.2}$} & {\tiny67.8$^{2.7}$} & {\tiny24.2$^{1.1}$} & {\color{red}{\tiny31.8$^{0.4}$}} & {\tiny34.2$^{0.9}$} & {\color{red}{\tiny36.5$^{1.3}$}}\\
{\tiny DHN+$\lambda$} & \textbf{{\tiny81.2$^{0.3}$}} & {\tiny81.1$^{0.6}$} & \textbf{{\tiny81.1$^{0.3}$}} & \textbf{{\tiny81.3$^{0.3}$}} & {\tiny80.6$^{0.4}$} & {\tiny81.3$^{0.2}$} & {\tiny81.6$^{0.2}$} & {\tiny81.7$^{0.2}$} & \textbf{{\tiny66.8$^{1.4}$}} & {\tiny67.3$^{1.0}$} & \textbf{{\tiny69.2$^{1.4}$}} & \textbf{{\tiny69.4$^{1.1}$}} & {\tiny25.1$^{0.6}$} & {\tiny32.4$^{0.4}$} & {\tiny35.7$^{1.2}$} & {\tiny38.2$^{0.3}$}\\
{\tiny DHN+HSWD} & {\tiny81.2$^{0.7}$} & \textbf{{\tiny81.4$^{0.5}$}} & {\tiny81.0$^{0.7}$} & {\tiny80.4$^{0.4}$} & \textbf{{\tiny80.7$^{0.1}$}} & {\tiny81.4$^{0.3}$} & \textbf{{\tiny81.9$^{0.2}$}} & {\tiny81.8$^{0.0}$} & {\tiny63.8$^{1.0}$} & \textbf{{\tiny67.9$^{1.7}$}} & {\tiny67.3$^{1.5}$} & {\tiny68.9$^{1.2}$} & \textbf{{\tiny33.1$^{0.7}$}} & \textbf{{\tiny39.9$^{0.6}$}} & \textbf{{\tiny42.8$^{0.3}$}} & \textbf{{\tiny44.4$^{0.5}$}}\\
{\tiny DHN+H²Q($L_2$)} & {\tiny80.5$^{0.5}$} & {\tiny80.7$^{0.6}$} & {\tiny79.9$^{0.7}$} & {\tiny80.5$^{0.6}$} & {\tiny80.5$^{0.3}$} & {\tiny81.4$^{0.2}$} & {\tiny81.5$^{0.2}$} & {\tiny82.0$^{0.2}$} & {\tiny64.6$^{1.4}$} & {\tiny67.2$^{1.4}$} & {\tiny68.0$^{3.1}$} & {\tiny68.9$^{2.7}$} & {\tiny26.0$^{1.3}$} & {\tiny34.4$^{0.4}$} & {\tiny36.4$^{1.1}$} & {\tiny38.8$^{1.5}$}\\
{\tiny DHN+H²Q($L_1$)} & {\tiny80.6$^{0.5}$} & {\tiny80.7$^{0.5}$} & {\tiny80.1$^{0.7}$} & {\tiny80.5$^{0.5}$} & {\tiny80.3$^{0.4}$} & {\tiny81.2$^{0.2}$} & {\tiny81.4$^{0.2}$} & {\tiny81.8$^{0.1}$} & {\tiny64.2$^{1.4}$} & {\tiny66.9$^{1.5}$} & {\tiny67.8$^{3.0}$} & {\tiny68.8$^{2.7}$} & {\tiny25.8$^{1.2}$} & {\tiny34.3$^{0.3}$} & {\tiny36.4$^{1.1}$} & {\tiny38.8$^{1.7}$}\\
{\tiny DHN+H²Q(min)} & {\tiny80.6$^{0.8}$} & {\tiny80.8$^{0.4}$} & {\tiny80.0$^{0.8}$} & {\tiny80.4$^{0.5}$} & {\tiny80.5$^{0.3}$} & {\tiny81.4$^{0.2}$} & {\tiny81.5$^{0.2}$} & {\tiny81.9$^{0.3}$} & {\tiny64.5$^{1.4}$} & {\tiny67.2$^{1.5}$} & {\tiny68.1$^{3.1}$} & {\tiny69.0$^{2.7}$} & {\tiny25.9$^{1.3}$} & {\tiny34.4$^{0.2}$} & {\tiny36.4$^{1.0}$} & {\tiny38.9$^{1.7}$}\\
{\tiny DHN+H²Q(bit)} & {\tiny80.7$^{0.6}$} & {\tiny80.8$^{0.4}$} & {\tiny80.1$^{0.8}$} & {\tiny80.5$^{0.4}$} & {\tiny80.5$^{0.3}$} & \textbf{{\tiny81.4$^{0.1}$}} & {\tiny81.7$^{0.2}$} & \textbf{{\tiny82.0$^{0.1}$}} & {\tiny64.4$^{1.4}$} & {\tiny67.1$^{1.6}$} & {\tiny68.1$^{3.2}$} & {\tiny69.0$^{2.7}$} & {\tiny25.6$^{1.1}$} & {\tiny34.3$^{0.4}$} & {\tiny36.2$^{0.8}$} & {\tiny38.5$^{1.7}$}\\[.2em]
{\tiny DPSH($\lambda=0$)} & {\tiny78.9$^{1.1}$} & {\tiny79.5$^{0.5}$} & {\tiny78.7$^{0.6}$} & {\tiny79.4$^{0.4}$} & {\tiny79.6$^{0.2}$} & {\tiny80.4$^{0.3}$} & {\tiny80.9$^{0.3}$} & {\tiny81.3$^{0.2}$} & {\tiny62.9$^{1.5}$} & {\tiny65.5$^{1.4}$} & {\tiny66.5$^{3.1}$} & {\tiny67.4$^{2.5}$} & {\tiny24.1$^{0.6}$} & {\tiny31.8$^{0.5}$} & {\tiny34.2$^{0.7}$} & {\tiny36.7$^{1.3}$}\\
{\tiny DPSH+ITQ} & {\tiny79.1$^{1.0}$} & {\color{red}{\tiny79.2$^{0.6}$}} & {\tiny78.7$^{1.0}$} & {\color{red}{\tiny79.3$^{0.4}$}} & {\color{red}{\tiny79.6$^{0.1}$}} & {\color{red}{\tiny80.4$^{0.1}$}} & {\tiny81.0$^{0.2}$} & {\tiny81.5$^{0.2}$} & {\tiny63.1$^{1.1}$} & {\tiny65.5$^{1.5}$} & {\tiny66.6$^{3.2}$} & {\tiny67.8$^{2.7}$} & {\tiny24.2$^{1.1}$} & {\color{red}{\tiny31.8$^{0.4}$}} & {\tiny34.2$^{0.9}$} & {\color{red}{\tiny36.5$^{1.3}$}}\\
{\tiny DPSH+$\lambda$} & \textbf{{\tiny81.2$^{0.7}$}} & {\tiny81.2$^{0.2}$} & \textbf{{\tiny81.5$^{0.5}$}} & \textbf{{\tiny81.1$^{0.4}$}} & \textbf{{\tiny81.0$^{0.3}$}} & \textbf{{\tiny81.9$^{0.1}$}} & \textbf{{\tiny82.1$^{0.3}$}} & \textbf{{\tiny82.1$^{0.4}$}} & \textbf{{\tiny68.0$^{1.6}$}} & \textbf{{\tiny71.2$^{0.1}$}} & \textbf{{\tiny71.6$^{1.3}$}} & \textbf{{\tiny72.4$^{0.2}$}} & \textbf{{\tiny36.5$^{1.7}$}} & \textbf{{\tiny42.2$^{1.7}$}} & \textbf{{\tiny46.0$^{1.0}$}} & \textbf{{\tiny49.9$^{0.3}$}}\\
{\tiny DPSH+HSWD} & {\tiny81.2$^{0.7}$} & \textbf{{\tiny81.4$^{0.5}$}} & {\tiny81.0$^{0.7}$} & {\tiny80.4$^{0.4}$} & {\tiny80.7$^{0.1}$} & {\tiny81.4$^{0.3}$} & {\tiny81.9$^{0.2}$} & {\tiny81.8$^{0.0}$} & {\tiny63.8$^{1.0}$} & {\tiny67.9$^{1.7}$} & {\tiny67.3$^{1.5}$} & {\tiny68.9$^{1.2}$} & {\tiny33.1$^{0.7}$} & {\tiny39.9$^{0.6}$} & {\tiny42.8$^{0.3}$} & {\tiny44.4$^{0.5}$}\\
{\tiny DPSH+H²Q($L_2$)} & {\tiny80.5$^{0.5}$} & {\tiny80.7$^{0.6}$} & {\tiny79.9$^{0.7}$} & {\tiny80.5$^{0.6}$} & {\tiny80.5$^{0.3}$} & {\tiny81.4$^{0.2}$} & {\tiny81.5$^{0.2}$} & {\tiny82.0$^{0.2}$} & {\tiny64.6$^{1.4}$} & {\tiny67.2$^{1.4}$} & {\tiny68.0$^{3.1}$} & {\tiny68.9$^{2.7}$} & {\tiny26.0$^{1.3}$} & {\tiny34.4$^{0.4}$} & {\tiny36.4$^{1.1}$} & {\tiny38.8$^{1.5}$}\\
{\tiny DPSH+H²Q($L_1$)} & {\tiny80.6$^{0.5}$} & {\tiny80.7$^{0.5}$} & {\tiny80.1$^{0.7}$} & {\tiny80.5$^{0.5}$} & {\tiny80.3$^{0.4}$} & {\tiny81.2$^{0.2}$} & {\tiny81.4$^{0.2}$} & {\tiny81.8$^{0.1}$} & {\tiny64.2$^{1.4}$} & {\tiny66.9$^{1.5}$} & {\tiny67.8$^{3.0}$} & {\tiny68.8$^{2.7}$} & {\tiny25.8$^{1.2}$} & {\tiny34.3$^{0.3}$} & {\tiny36.4$^{1.1}$} & {\tiny38.8$^{1.7}$}\\
{\tiny DPSH+H²Q(min)} & {\tiny80.6$^{0.8}$} & {\tiny80.8$^{0.4}$} & {\tiny80.0$^{0.8}$} & {\tiny80.4$^{0.5}$} & {\tiny80.5$^{0.3}$} & {\tiny81.4$^{0.2}$} & {\tiny81.5$^{0.2}$} & {\tiny81.9$^{0.3}$} & {\tiny64.5$^{1.4}$} & {\tiny67.2$^{1.5}$} & {\tiny68.1$^{3.1}$} & {\tiny69.0$^{2.7}$} & {\tiny25.9$^{1.3}$} & {\tiny34.4$^{0.2}$} & {\tiny36.4$^{1.0}$} & {\tiny38.9$^{1.7}$}\\
{\tiny DPSH+H²Q(bit)} & {\tiny80.7$^{0.6}$} & {\tiny80.8$^{0.4}$} & {\tiny80.1$^{0.8}$} & {\tiny80.5$^{0.4}$} & {\tiny80.5$^{0.3}$} & {\tiny81.4$^{0.1}$} & {\tiny81.7$^{0.2}$} & {\tiny82.0$^{0.1}$} & {\tiny64.4$^{1.4}$} & {\tiny67.1$^{1.6}$} & {\tiny68.1$^{3.2}$} & {\tiny69.0$^{2.7}$} & {\tiny25.6$^{1.1}$} & {\tiny34.3$^{0.4}$} & {\tiny36.2$^{0.8}$} & {\tiny38.5$^{1.7}$}\\[.2em]
{\tiny WGL.($\lambda=0$)} & {\tiny78.3$^{0.5}$} & {\tiny76.9$^{0.5}$} & {\tiny75.6$^{0.7}$} & {\tiny76.1$^{0.5}$} & {\tiny79.4$^{0.6}$} & {\tiny79.6$^{0.4}$} & {\tiny79.4$^{0.5}$} & {\tiny78.8$^{0.3}$} & {\tiny64.4$^{1.2}$} & {\tiny64.0$^{0.4}$} & {\tiny64.0$^{0.2}$} & {\tiny64.0$^{0.3}$} & {\tiny49.8$^{1.9}$} & {\tiny46.5$^{0.6}$} & {\tiny47.6$^{2.2}$} & {\tiny48.4$^{1.8}$}\\
{\tiny WGL.+ITQ} & {\color{red}{\tiny77.9$^{1.0}$}} & {\tiny77.6$^{0.6}$} & {\tiny76.1$^{1.1}$} & {\tiny76.2$^{1.2}$} & {\color{red}{\tiny79.3$^{0.4}$}} & {\tiny79.7$^{0.5}$} & {\tiny79.4$^{0.6}$} & {\tiny79.1$^{0.4}$} & {\color{red}{\tiny63.9$^{0.7}$}} & {\tiny64.1$^{0.3}$} & {\tiny64.1$^{0.2}$} & {\tiny64.7$^{0.3}$} & {\color{red}{\tiny49.5$^{1.7}$}} & {\tiny47.4$^{0.9}$} & {\tiny48.1$^{2.5}$} & {\tiny49.2$^{1.8}$}\\
{\tiny WGL.+$\lambda$} & {\tiny79.6$^{0.8}$} & {\tiny80.0$^{0.2}$} & {\tiny80.2$^{0.2}$} & {\tiny79.4$^{1.7}$} & {\tiny79.9$^{0.2}$} & {\tiny80.7$^{0.5}$} & {\tiny80.1$^{0.4}$} & {\tiny80.5$^{0.7}$} & \textbf{{\tiny66.3$^{1.9}$}} & {\tiny67.0$^{1.4}$} & \textbf{{\tiny67.7$^{1.4}$}} & \textbf{{\tiny67.2$^{1.4}$}} & {\tiny55.3$^{1.3}$} & \textbf{{\tiny57.1$^{1.1}$}} & \textbf{{\tiny57.0$^{0.9}$}} & \textbf{{\tiny56.8$^{0.7}$}}\\
{\tiny WGL.+HSWD} & {\tiny79.0$^{0.5}$} & {\tiny79.2$^{0.8}$} & {\tiny77.7$^{0.6}$} & {\tiny76.4$^{0.8}$} & {\tiny79.7$^{0.3}$} & {\color{red}{\tiny79.0$^{0.2}$}} & {\color{red}{\tiny78.3$^{0.8}$}} & {\color{red}{\tiny78.5$^{0.2}$}} & {\tiny65.1$^{2.2}$} & \textbf{{\tiny67.2$^{0.7}$}} & {\tiny65.6$^{2.1}$} & {\tiny65.6$^{0.3}$} & \textbf{{\tiny55.3$^{1.4}$}} & {\tiny52.0$^{0.9}$} & {\color{red}{\tiny47.4$^{1.3}$}} & {\color{red}{\tiny47.2$^{2.1}$}}\\
{\tiny WGL.+H²Q($L_2$)} & {\tiny81.6$^{0.3}$} & {\tiny80.9$^{0.6}$} & {\tiny80.5$^{0.2}$} & {\tiny80.2$^{0.2}$} & {\tiny81.0$^{0.3}$} & {\tiny81.7$^{0.3}$} & {\tiny81.2$^{0.2}$} & {\tiny81.4$^{0.2}$} & {\tiny66.2$^{0.9}$} & {\tiny66.5$^{0.4}$} & {\tiny66.8$^{0.3}$} & {\tiny66.7$^{0.2}$} & {\tiny54.4$^{1.5}$} & {\tiny52.8$^{0.8}$} & {\tiny53.9$^{1.9}$} & {\tiny54.7$^{1.4}$}\\
{\tiny WGL.+H²Q($L_1$)} & {\tiny81.9$^{0.3}$} & \textbf{{\tiny81.1$^{0.6}$}} & {\tiny80.6$^{0.3}$} & {\tiny80.3$^{0.6}$} & \textbf{{\tiny81.1$^{0.3}$}} & {\tiny81.6$^{0.2}$} & {\tiny81.3$^{0.2}$} & {\tiny81.2$^{0.3}$} & {\tiny66.1$^{0.8}$} & {\tiny66.4$^{0.5}$} & {\tiny66.8$^{0.3}$} & {\tiny66.6$^{0.3}$} & {\tiny54.5$^{1.5}$} & {\tiny52.5$^{0.4}$} & {\tiny53.6$^{2.1}$} & {\tiny53.9$^{1.5}$}\\
{\tiny WGL.+H²Q(min)} & \textbf{{\tiny82.0$^{0.7}$}} & {\tiny81.0$^{0.6}$} & {\tiny81.0$^{0.4}$} & {\tiny80.7$^{0.2}$} & {\tiny80.9$^{0.3}$} & \textbf{{\tiny81.7$^{0.2}$}} & \textbf{{\tiny81.4$^{0.1}$}} & {\tiny81.3$^{0.3}$} & {\tiny66.2$^{0.7}$} & {\tiny66.5$^{0.4}$} & {\tiny66.8$^{0.3}$} & {\tiny66.7$^{0.2}$} & {\tiny54.1$^{1.1}$} & {\tiny52.7$^{0.7}$} & {\tiny53.7$^{1.5}$} & {\tiny54.1$^{1.8}$}\\
{\tiny WGL.+H²Q(bit)} & {\tiny81.9$^{0.5}$} & {\tiny81.1$^{0.5}$} & \textbf{{\tiny81.0$^{0.3}$}} & \textbf{{\tiny80.7$^{0.4}$}} & {\tiny81.0$^{0.4}$} & {\tiny81.7$^{0.3}$} & {\tiny81.3$^{0.3}$} & \textbf{{\tiny81.4$^{0.3}$}} & {\tiny66.2$^{1.0}$} & {\tiny66.3$^{0.3}$} & {\tiny66.6$^{0.3}$} & {\tiny66.5$^{0.3}$} & {\tiny53.8$^{1.6}$} & {\tiny52.0$^{0.8}$} & {\tiny53.2$^{1.8}$} & {\tiny53.9$^{1.2}$}\\[.2em]
{\tiny HyP²($\lambda=0$)} & {\tiny80.5$^{0.8}$} & {\tiny81.1$^{0.5}$} & {\tiny81.7$^{0.3}$} & {\tiny81.8$^{0.5}$} & {\tiny81.9$^{0.1}$} & {\tiny82.5$^{0.2}$} & {\tiny83.1$^{0.1}$} & {\tiny83.0$^{0.2}$} & {\tiny71.9$^{0.4}$} & {\tiny74.1$^{0.4}$} & {\tiny74.8$^{0.2}$} & {\tiny74.9$^{0.6}$} & {\tiny54.1$^{1.4}$} & {\tiny56.9$^{0.4}$} & {\tiny57.7$^{0.5}$} & {\tiny56.5$^{0.2}$}\\
{\tiny HyP²+ITQ} & {\color{red}{\tiny80.2$^{0.2}$}} & {\color{red}{\tiny80.7$^{0.7}$}} & {\color{red}{\tiny81.4$^{0.2}$}} & {\color{red}{\tiny81.8$^{0.5}$}} & {\color{red}{\tiny81.6$^{0.3}$}} & {\tiny82.5$^{0.2}$} & {\color{red}{\tiny83.0$^{0.1}$}} & {\color{red}{\tiny83.0$^{0.2}$}} & {\color{red}{\tiny71.6$^{0.7}$}} & {\color{red}{\tiny73.9$^{0.2}$}} & {\color{red}{\tiny74.8$^{0.2}$}} & {\color{red}{\tiny74.8$^{0.6}$}} & {\color{red}{\tiny53.5$^{1.5}$}} & {\color{red}{\tiny56.7$^{0.5}$}} & {\color{red}{\tiny57.6$^{0.4}$}} & {\color{red}{\tiny56.3$^{0.5}$}}\\
{\tiny HyP²+$\lambda$} & {\tiny82.1$^{0.3}$} & {\tiny82.2$^{0.1}$} & {\tiny82.2$^{0.3}$} & {\tiny82.1$^{0.3}$} & {\tiny82.1$^{0.1}$} & {\tiny82.5$^{0.2}$} & {\color{red}{\tiny82.7$^{0.2}$}} & {\color{red}{\tiny82.5$^{0.2}$}} & {\tiny73.0$^{0.4}$} & {\color{red}{\tiny73.6$^{0.4}$}} & {\color{red}{\tiny72.6$^{0.6}$}} & {\color{red}{\tiny71.4$^{0.6}$}} & \textbf{{\tiny58.4$^{0.6}$}} & \textbf{{\tiny62.2$^{0.7}$}} & \textbf{{\tiny62.4$^{0.2}$}} & \textbf{{\tiny61.3$^{0.8}$}}\\
{\tiny HyP²+HSWD} & {\tiny82.1$^{0.6}$} & {\tiny82.3$^{0.2}$} & {\tiny82.4$^{0.1}$} & {\tiny82.0$^{0.4}$} & {\tiny82.0$^{0.1}$} & {\tiny82.6$^{0.2}$} & {\tiny83.2$^{0.1}$} & {\tiny83.1$^{0.1}$} & {\tiny72.1$^{0.4}$} & {\tiny74.7$^{0.4}$} & {\tiny74.9$^{0.2}$} & {\tiny74.9$^{0.2}$} & {\tiny56.8$^{0.9}$} & {\tiny60.2$^{0.5}$} & {\tiny60.8$^{0.5}$} & {\tiny58.8$^{0.7}$}\\
{\tiny HyP²+H²Q($L_2$)} & \textbf{{\tiny82.3$^{0.3}$}} & {\tiny82.5$^{0.5}$} & {\tiny82.9$^{0.2}$} & {\tiny83.1$^{0.3}$} & {\tiny82.5$^{0.2}$} & \textbf{{\tiny83.2$^{0.1}$}} & \textbf{{\tiny83.4$^{0.1}$}} & \textbf{{\tiny83.3$^{0.1}$}} & {\tiny73.9$^{0.2}$} & \textbf{{\tiny75.4$^{0.2}$}} & {\tiny75.9$^{0.3}$} & {\tiny75.7$^{0.3}$} & {\tiny57.3$^{0.9}$} & {\tiny60.7$^{0.4}$} & {\tiny61.5$^{0.3}$} & {\tiny60.6$^{0.4}$}\\
{\tiny HyP²+H²Q($L_1$)} & {\tiny82.3$^{0.2}$} & {\tiny82.6$^{0.5}$} & \textbf{{\tiny83.0$^{0.3}$}} & {\tiny83.1$^{0.4}$} & {\tiny82.5$^{0.2}$} & {\tiny83.1$^{0.1}$} & {\tiny83.3$^{0.1}$} & {\tiny83.3$^{0.1}$} & {\tiny73.8$^{0.3}$} & {\tiny75.3$^{0.2}$} & {\tiny75.8$^{0.3}$} & {\tiny75.6$^{0.4}$} & {\tiny57.0$^{0.8}$} & {\tiny60.5$^{0.4}$} & {\tiny61.2$^{0.2}$} & {\tiny60.7$^{0.4}$}\\
{\tiny HyP²+H²Q(min)} & {\tiny82.3$^{0.2}$} & {\tiny82.7$^{0.5}$} & {\tiny82.8$^{0.2}$} & \textbf{{\tiny83.1$^{0.3}$}} & \textbf{{\tiny82.6$^{0.2}$}} & {\tiny83.2$^{0.2}$} & {\tiny83.4$^{0.1}$} & {\tiny83.3$^{0.2}$} & \textbf{{\tiny73.9$^{0.3}$}} & {\tiny75.3$^{0.2}$} & {\tiny75.9$^{0.2}$} & \textbf{{\tiny75.7$^{0.3}$}} & {\tiny57.1$^{0.9}$} & {\tiny60.3$^{0.3}$} & {\tiny61.3$^{0.4}$} & {\tiny60.6$^{0.3}$}\\
{\tiny HyP²+H²Q(bit)} & {\tiny82.3$^{0.2}$} & \textbf{{\tiny82.7$^{0.4}$}} & {\tiny82.9$^{0.2}$} & {\tiny83.0$^{0.4}$} & {\tiny82.5$^{0.2}$} & {\tiny83.2$^{0.1}$} & {\tiny83.4$^{0.1}$} & {\tiny83.3$^{0.2}$} & {\tiny73.7$^{0.2}$} & {\tiny75.3$^{0.2}$} & \textbf{{\tiny75.9$^{0.3}$}} & {\tiny75.6$^{0.3}$} & {\tiny56.8$^{0.8}$} & {\tiny60.0$^{0.4}$} & {\tiny60.9$^{0.1}$} & {\tiny60.2$^{0.3}$}\\[.2em]
\end{tabular}
\caption{\footnotesize{Full comparison of \texttt{mAP@k} before (level $\lambda = 0$) and after using each quantization strategy on AlexNet. Performance metrics in red indicate a decrease in performance after the quantization strategy and performance metrics in bold indicate which quantization strategy gave the best overall metric. The small superscript numbers indicate the standard deviation of the metrics.}}
\label{sm-tab:full_CNNF_alexnet_SM}
\end{table*}

\begin{table*}[]\centering
\vspace*{-1cm}
\setlength\extrarowheight{-2pt}
\begin{tabular}{l@{\hskip .01in}|c@{\hskip .01in}c@{\hskip .01in}c@{\hskip .01in}c|c@{\hskip .01in}c@{\hskip .01in}c@{\hskip .01in}c|c@{\hskip .01in}c@{\hskip .01in}c@{\hskip .01in}c|c@{\hskip .01in}c@{\hskip .01in}c@{\hskip .01in}c}
& \multicolumn{4}{c|}{CIFAR 10} & \multicolumn{4}{c|}{NUS WIDE} & \multicolumn{4}{c|}{MS COCO} & \multicolumn{4}{c}{ImageNet} \\ \hline
n bits & 16 & 32 & 48 & 64 & 16 & 32 & 48 & 64 & 16 & 32 & 48 & 64 & 16 & 32 & 48 & 64\\ \hline
{\tiny CEL($\lambda=0$)} & {\tiny85.1$^{0.3}$} & {\tiny85.6$^{0.4}$} & {\tiny85.8$^{0.3}$} & {\tiny85.7$^{0.4}$} & {\tiny82.7$^{0.3}$} & {\tiny83.1$^{0.5}$} & {\tiny83.3$^{0.2}$} & {\tiny82.8$^{0.4}$} & {\tiny73.3$^{0.7}$} & {\tiny75.7$^{0.7}$} & {\tiny76.2$^{0.4}$} & {\tiny76.4$^{0.2}$} & {\tiny69.9$^{0.7}$} & {\tiny72.8$^{0.3}$} & {\tiny74.4$^{0.4}$} & {\tiny72.5$^{0.3}$}\\
{\tiny CEL+ITQ} & {\tiny85.2$^{0.3}$} & {\tiny85.7$^{0.1}$} & {\color{red}{\tiny85.5$^{0.2}$}} & {\color{red}{\tiny85.6$^{0.4}$}} & {\color{red}{\tiny82.5$^{0.4}$}} & {\tiny83.2$^{0.3}$} & {\color{red}{\tiny83.3$^{0.1}$}} & {\tiny83.0$^{0.5}$} & {\tiny73.5$^{0.4}$} & {\tiny75.7$^{0.9}$} & {\tiny76.3$^{0.5}$} & {\color{red}{\tiny76.4$^{0.5}$}} & {\color{red}{\tiny69.3$^{0.5}$}} & {\tiny73.0$^{0.3}$} & {\color{red}{\tiny74.2$^{0.7}$}} & {\tiny72.7$^{0.1}$}\\
{\tiny CEL+$\lambda$} & {\color{red}{\tiny84.8$^{0.7}$}} & {\color{red}{\tiny85.3$^{0.2}$}} & {\color{red}{\tiny82.9$^{1.2}$}} & {\color{red}{\tiny64.6$^{15.5}$}} & {\color{red}{\tiny81.2$^{0.3}$}} & {\color{red}{\tiny81.5$^{0.1}$}} & {\color{red}{\tiny81.3$^{0.2}$}} & {\color{red}{\tiny81.2$^{0.1}$}} & {\color{red}{\tiny73.0$^{0.8}$}} & {\color{red}{\tiny73.3$^{0.7}$}} & {\color{red}{\tiny73.0$^{0.4}$}} & {\color{red}{\tiny72.5$^{0.3}$}} & {\color{red}{\tiny19.1$^{1.8}$}} & {\color{red}{\tiny21.7$^{3.0}$}} & {\color{red}{\tiny25.5$^{2.3}$}} & {\color{red}{\tiny29.3$^{2.7}$}}\\
{\tiny CEL+HSWD} & \textbf{{\tiny86.6$^{0.3}$}} & \textbf{{\tiny86.7$^{0.3}$}} & \textbf{{\tiny86.9$^{0.2}$}} & \textbf{{\tiny87.1$^{0.3}$}} & {\color{red}{\tiny81.6$^{0.5}$}} & {\color{red}{\tiny82.4$^{0.2}$}} & {\color{red}{\tiny82.4$^{0.8}$}} & {\color{red}{\tiny82.5$^{0.6}$}} & {\color{red}{\tiny72.7$^{0.3}$}} & {\tiny75.7$^{1.0}$} & {\color{red}{\tiny76.0$^{0.4}$}} & {\color{red}{\tiny75.7$^{0.9}$}} & {\color{red}{\tiny69.3$^{0.9}$}} & {\tiny72.8$^{0.9}$} & {\color{red}{\tiny73.6$^{1.1}$}} & {\color{red}{\tiny71.8$^{0.2}$}}\\
{\tiny CEL+H²Q($L_2$)} & {\tiny86.1$^{0.1}$} & {\tiny86.5$^{0.4}$} & {\tiny86.8$^{0.1}$} & {\tiny86.6$^{0.5}$} & {\tiny83.4$^{0.2}$} & {\tiny84.0$^{0.3}$} & {\tiny84.3$^{0.1}$} & \textbf{{\tiny84.1$^{0.2}$}} & {\tiny76.1$^{0.3}$} & {\tiny78.0$^{0.5}$} & {\tiny78.5$^{0.4}$} & {\tiny78.5$^{0.4}$} & \textbf{{\tiny72.8$^{0.8}$}} & \textbf{{\tiny75.8$^{0.2}$}} & \textbf{{\tiny77.3$^{0.5}$}} & \textbf{{\tiny75.8$^{0.4}$}}\\
{\tiny CEL+H²Q($L_1$)} & {\tiny86.1$^{0.2}$} & {\tiny86.6$^{0.3}$} & {\tiny86.8$^{0.2}$} & {\tiny86.5$^{0.4}$} & {\tiny83.4$^{0.2}$} & {\tiny84.0$^{0.3}$} & \textbf{{\tiny84.3$^{0.1}$}} & {\tiny84.0$^{0.2}$} & {\tiny76.1$^{0.4}$} & {\tiny78.0$^{0.6}$} & \textbf{{\tiny78.5$^{0.5}$}} & {\tiny78.5$^{0.4}$} & {\tiny72.4$^{1.0}$} & {\tiny75.7$^{0.3}$} & {\tiny77.1$^{0.6}$} & {\tiny75.8$^{0.3}$}\\
{\tiny CEL+H²Q(min)} & {\tiny86.1$^{0.1}$} & {\tiny86.6$^{0.2}$} & {\tiny86.8$^{0.1}$} & {\tiny86.6$^{0.3}$} & {\tiny83.3$^{0.3}$} & {\tiny84.1$^{0.3}$} & {\tiny84.3$^{0.1}$} & {\tiny84.1$^{0.2}$} & \textbf{{\tiny76.2$^{0.4}$}} & {\tiny78.0$^{0.7}$} & {\tiny78.5$^{0.5}$} & \textbf{{\tiny78.5$^{0.4}$}} & {\tiny72.4$^{1.0}$} & {\tiny75.5$^{0.4}$} & {\tiny77.0$^{0.4}$} & {\tiny75.5$^{0.3}$}\\
{\tiny CEL+H²Q(bit)} & {\tiny86.3$^{0.2}$} & {\tiny86.5$^{0.2}$} & {\tiny86.8$^{0.1}$} & {\tiny86.7$^{0.3}$} & \textbf{{\tiny83.4$^{0.3}$}} & \textbf{{\tiny84.1$^{0.3}$}} & {\tiny84.3$^{0.1}$} & {\tiny84.1$^{0.2}$} & {\tiny76.2$^{0.4}$} & \textbf{{\tiny78.1$^{0.6}$}} & {\tiny78.5$^{0.5}$} & {\tiny78.5$^{0.4}$} & {\tiny72.3$^{1.1}$} & {\tiny75.4$^{0.3}$} & {\tiny77.1$^{0.4}$} & {\tiny75.4$^{0.4}$}\\[.2em]
{\tiny DCH($\lambda=0$)} & {\tiny84.5$^{0.4}$} & {\tiny82.6$^{1.2}$} & {\tiny80.1$^{1.0}$} & {\tiny80.2$^{0.2}$} & {\tiny81.9$^{0.6}$} & {\tiny81.3$^{0.1}$} & {\tiny80.5$^{0.2}$} & {\tiny80.0$^{0.3}$} & {\tiny73.5$^{0.8}$} & {\tiny69.4$^{1.7}$} & {\tiny70.0$^{1.6}$} & {\tiny68.4$^{0.7}$} & {\tiny77.5$^{0.3}$} & {\tiny75.6$^{1.2}$} & {\tiny72.2$^{1.1}$} & {\tiny69.6$^{1.2}$}\\
{\tiny DCH+ITQ} & {\color{red}{\tiny84.2$^{0.3}$}} & {\color{red}{\tiny82.5$^{1.0}$}} & {\tiny80.1$^{0.8}$} & {\color{red}{\tiny80.2$^{0.6}$}} & {\color{red}{\tiny81.9$^{0.6}$}} & {\tiny81.4$^{0.1}$} & {\tiny80.6$^{0.4}$} & {\tiny80.2$^{0.2}$} & {\tiny73.6$^{0.5}$} & {\tiny70.0$^{1.4}$} & {\tiny70.8$^{1.4}$} & {\tiny69.4$^{0.4}$} & {\color{red}{\tiny77.1$^{0.4}$}} & {\color{red}{\tiny75.5$^{0.8}$}} & {\tiny72.6$^{1.3}$} & {\tiny70.3$^{0.7}$}\\
{\tiny DCH+$\lambda$} & {\color{red}{\tiny84.0$^{0.6}$}} & {\tiny82.9$^{0.6}$} & \textbf{{\tiny82.8$^{0.9}$}} & \textbf{{\tiny82.9$^{0.2}$}} & {\tiny81.9$^{0.2}$} & {\tiny81.8$^{0.4}$} & {\tiny81.1$^{0.2}$} & \textbf{{\tiny81.2$^{0.2}$}} & {\tiny75.7$^{0.8}$} & \textbf{{\tiny76.0$^{1.2}$}} & \textbf{{\tiny76.7$^{1.2}$}} & \textbf{{\tiny76.0$^{1.2}$}} & \textbf{{\tiny79.8$^{0.3}$}} & \textbf{{\tiny80.2$^{0.5}$}} & \textbf{{\tiny79.8$^{0.3}$}} & \textbf{{\tiny80.1$^{0.5}$}}\\
{\tiny DCH+HSWD} & {\tiny85.4$^{0.5}$} & {\tiny83.2$^{0.6}$} & {\tiny81.9$^{1.7}$} & {\tiny81.3$^{1.2}$} & {\color{red}{\tiny81.5$^{0.2}$}} & {\color{red}{\tiny80.5$^{0.2}$}} & {\color{red}{\tiny79.7$^{0.2}$}} & {\color{red}{\tiny79.1$^{0.1}$}} & {\color{red}{\tiny72.3$^{1.3}$}} & {\tiny69.7$^{3.5}$} & {\color{red}{\tiny69.1$^{1.3}$}} & {\color{red}{\tiny67.0$^{0.7}$}} & {\tiny78.5$^{0.6}$} & {\color{red}{\tiny75.2$^{1.0}$}} & {\color{red}{\tiny71.6$^{1.6}$}} & {\tiny71.3$^{1.4}$}\\
{\tiny DCH+H²Q($L_2$)} & {\tiny85.6$^{0.2}$} & {\tiny83.9$^{0.8}$} & {\tiny81.9$^{0.5}$} & {\tiny81.8$^{0.5}$} & {\tiny82.6$^{0.3}$} & {\tiny82.3$^{0.1}$} & {\tiny81.5$^{0.2}$} & {\tiny80.9$^{0.3}$} & {\tiny76.5$^{0.6}$} & {\tiny72.6$^{1.7}$} & {\tiny72.9$^{1.3}$} & {\tiny71.3$^{0.4}$} & {\tiny79.6$^{0.6}$} & {\tiny78.0$^{0.7}$} & {\tiny74.7$^{0.8}$} & {\tiny72.4$^{0.8}$}\\
{\tiny DCH+H²Q($L_1$)} & {\tiny85.6$^{0.2}$} & {\tiny84.0$^{0.7}$} & {\tiny81.8$^{0.8}$} & {\tiny81.5$^{0.5}$} & {\tiny82.5$^{0.5}$} & {\tiny82.2$^{0.2}$} & {\tiny81.5$^{0.1}$} & {\tiny80.8$^{0.3}$} & {\tiny76.5$^{0.6}$} & {\tiny72.6$^{1.6}$} & {\tiny72.8$^{1.4}$} & {\tiny71.3$^{0.3}$} & {\tiny79.2$^{0.8}$} & {\tiny77.4$^{0.5}$} & {\tiny74.0$^{0.8}$} & {\tiny71.5$^{0.4}$}\\
{\tiny DCH+H²Q(min)} & {\tiny85.6$^{0.3}$} & {\tiny84.0$^{0.8}$} & {\tiny81.8$^{0.6}$} & {\tiny81.8$^{0.6}$} & {\tiny82.7$^{0.4}$} & {\tiny82.3$^{0.2}$} & {\tiny81.5$^{0.1}$} & {\tiny80.9$^{0.4}$} & {\tiny76.5$^{0.5}$} & {\tiny72.7$^{1.7}$} & {\tiny72.9$^{1.3}$} & {\tiny71.3$^{0.4}$} & {\tiny79.1$^{0.6}$} & {\tiny77.0$^{0.7}$} & {\tiny73.3$^{0.6}$} & {\tiny70.7$^{0.5}$}\\
{\tiny DCH+H²Q(bit)} & \textbf{{\tiny85.7$^{0.2}$}} & \textbf{{\tiny84.1$^{0.8}$}} & {\tiny81.8$^{0.6}$} & {\tiny81.5$^{0.5}$} & \textbf{{\tiny82.7$^{0.4}$}} & \textbf{{\tiny82.4$^{0.1}$}} & \textbf{{\tiny81.6$^{0.1}$}} & {\tiny80.9$^{0.3}$} & \textbf{{\tiny76.6$^{0.6}$}} & {\tiny72.6$^{1.7}$} & {\tiny72.9$^{1.3}$} & {\tiny71.2$^{0.5}$} & {\tiny78.7$^{0.8}$} & {\tiny76.6$^{0.8}$} & {\tiny72.8$^{0.8}$} & {\tiny70.0$^{0.3}$}\\[.2em]
{\tiny DHN($\lambda=0$)} & {\tiny83.1$^{2.3}$} & {\tiny85.5$^{0.5}$} & {\tiny85.9$^{0.5}$} & {\tiny85.8$^{0.3}$} & {\tiny82.9$^{0.2}$} & {\tiny83.2$^{0.3}$} & {\tiny83.6$^{0.1}$} & {\tiny83.3$^{0.3}$} & {\tiny66.3$^{1.3}$} & {\tiny69.5$^{1.4}$} & {\tiny70.0$^{1.0}$} & {\tiny71.4$^{1.1}$} & {\tiny44.8$^{0.6}$} & {\tiny55.6$^{0.8}$} & {\tiny59.9$^{1.0}$} & {\tiny62.5$^{0.8}$}\\
{\tiny DHN+ITQ} & {\tiny83.6$^{2.1}$} & {\color{red}{\tiny85.5$^{0.5}$}} & {\color{red}{\tiny85.9$^{0.5}$}} & {\tiny85.9$^{0.4}$} & {\color{red}{\tiny82.8$^{0.3}$}} & {\color{red}{\tiny83.2$^{0.3}$}} & {\color{red}{\tiny83.6$^{0.1}$}} & {\tiny83.3$^{0.3}$} & {\tiny66.4$^{1.3}$} & {\tiny69.8$^{1.3}$} & {\tiny70.6$^{1.1}$} & {\tiny71.7$^{1.0}$} & {\color{red}{\tiny44.4$^{0.6}$}} & {\color{red}{\tiny55.1$^{0.5}$}} & {\color{red}{\tiny59.9$^{0.8}$}} & {\color{red}{\tiny62.5$^{0.5}$}}\\
{\tiny DHN+$\lambda$} & {\tiny86.2$^{0.4}$} & {\tiny86.6$^{0.2}$} & \textbf{{\tiny86.9$^{0.6}$}} & {\tiny86.6$^{0.3}$} & {\tiny83.2$^{0.3}$} & {\tiny83.6$^{0.2}$} & {\tiny83.6$^{0.2}$} & {\tiny83.7$^{0.1}$} & \textbf{{\tiny68.6$^{2.5}$}} & {\tiny70.7$^{1.2}$} & {\color{red}{\tiny69.9$^{0.7}$}} & {\tiny71.5$^{1.4}$} & {\tiny45.4$^{0.4}$} & {\color{red}{\tiny54.4$^{1.1}$}} & {\color{red}{\tiny59.1$^{0.9}$}} & {\color{red}{\tiny62.3$^{1.2}$}}\\
{\tiny DHN+HSWD} & \textbf{{\tiny86.3$^{0.5}$}} & \textbf{{\tiny86.9$^{0.3}$}} & {\tiny86.9$^{0.2}$} & {\tiny86.5$^{0.4}$} & {\tiny83.3$^{0.1}$} & {\tiny83.7$^{0.2}$} & {\tiny83.9$^{0.3}$} & \textbf{{\tiny84.1$^{0.3}$}} & {\tiny68.1$^{2.0}$} & {\tiny70.3$^{1.5}$} & {\tiny70.5$^{1.0}$} & {\tiny72.8$^{1.4}$} & \textbf{{\tiny56.0$^{0.7}$}} & \textbf{{\tiny63.9$^{0.9}$}} & \textbf{{\tiny66.5$^{0.7}$}} & \textbf{{\tiny68.1$^{0.8}$}}\\
{\tiny DHN+H²Q($L_2$)} & {\tiny84.2$^{1.8}$} & {\tiny86.2$^{0.4}$} & {\tiny86.8$^{0.6}$} & {\tiny86.8$^{0.4}$} & \textbf{{\tiny83.4$^{0.2}$}} & \textbf{{\tiny83.9$^{0.2}$}} & \textbf{{\tiny84.1$^{0.1}$}} & {\tiny83.9$^{0.2}$} & {\tiny68.2$^{1.5}$} & {\tiny71.7$^{1.2}$} & \textbf{{\tiny72.7$^{1.1}$}} & \textbf{{\tiny73.7$^{0.9}$}} & {\tiny49.1$^{1.4}$} & {\tiny59.9$^{0.2}$} & {\tiny64.1$^{1.0}$} & {\tiny66.1$^{0.5}$}\\
{\tiny DHN+H²Q($L_1$)} & {\tiny84.0$^{2.0}$} & {\tiny86.2$^{0.4}$} & {\tiny86.8$^{0.6}$} & {\tiny86.7$^{0.5}$} & {\tiny83.2$^{0.4}$} & {\tiny83.8$^{0.2}$} & {\tiny84.0$^{0.1}$} & {\tiny83.7$^{0.2}$} & {\tiny68.1$^{1.5}$} & {\tiny71.7$^{1.4}$} & {\tiny72.6$^{1.0}$} & {\tiny73.6$^{0.8}$} & {\tiny49.2$^{1.3}$} & {\tiny59.6$^{0.2}$} & {\tiny63.9$^{1.1}$} & {\tiny65.8$^{0.6}$}\\
{\tiny DHN+H²Q(min)} & {\tiny84.1$^{1.8}$} & {\tiny86.2$^{0.3}$} & {\tiny86.8$^{0.6}$} & {\tiny86.7$^{0.4}$} & {\tiny83.4$^{0.2}$} & {\tiny83.8$^{0.2}$} & {\tiny84.0$^{0.1}$} & {\tiny83.8$^{0.3}$} & {\tiny68.0$^{1.4}$} & \textbf{{\tiny72.0$^{1.3}$}} & {\tiny72.6$^{1.1}$} & {\tiny73.7$^{0.9}$} & {\tiny49.0$^{1.1}$} & {\tiny59.1$^{0.6}$} & {\tiny63.7$^{1.1}$} & {\tiny66.1$^{0.7}$}\\
{\tiny DHN+H²Q(bit)} & {\tiny84.1$^{1.9}$} & {\tiny86.3$^{0.3}$} & {\tiny86.8$^{0.5}$} & \textbf{{\tiny86.8$^{0.5}$}} & {\tiny83.4$^{0.2}$} & {\tiny83.9$^{0.2}$} & {\tiny84.1$^{0.1}$} & {\tiny83.8$^{0.3}$} & {\tiny68.0$^{1.6}$} & {\tiny71.8$^{1.2}$} & {\tiny72.4$^{1.1}$} & {\tiny73.6$^{0.9}$} & {\tiny48.7$^{0.7}$} & {\tiny59.2$^{0.7}$} & {\tiny63.8$^{1.0}$} & {\tiny66.0$^{0.6}$}\\[.2em]
{\tiny DPSH($\lambda=0$)} & {\tiny83.1$^{2.3}$} & {\tiny85.5$^{0.5}$} & {\tiny85.9$^{0.5}$} & {\tiny85.8$^{0.3}$} & {\tiny82.9$^{0.2}$} & {\tiny83.2$^{0.3}$} & {\tiny83.6$^{0.1}$} & {\tiny83.3$^{0.3}$} & {\tiny66.3$^{1.3}$} & {\tiny69.5$^{1.4}$} & {\tiny70.0$^{1.0}$} & {\tiny71.4$^{1.1}$} & {\tiny44.8$^{0.6}$} & {\tiny55.6$^{0.8}$} & {\tiny59.9$^{1.0}$} & {\tiny62.5$^{0.8}$}\\
{\tiny DPSH+ITQ} & {\tiny83.6$^{2.1}$} & {\color{red}{\tiny85.5$^{0.5}$}} & {\color{red}{\tiny85.9$^{0.5}$}} & {\tiny85.9$^{0.4}$} & {\color{red}{\tiny82.8$^{0.3}$}} & {\color{red}{\tiny83.2$^{0.3}$}} & {\color{red}{\tiny83.6$^{0.1}$}} & {\tiny83.3$^{0.3}$} & {\tiny66.4$^{1.3}$} & {\tiny69.8$^{1.3}$} & {\tiny70.6$^{1.1}$} & {\tiny71.7$^{1.0}$} & {\color{red}{\tiny44.4$^{0.6}$}} & {\color{red}{\tiny55.1$^{0.5}$}} & {\color{red}{\tiny59.9$^{0.8}$}} & {\color{red}{\tiny62.5$^{0.5}$}}\\
{\tiny DPSH+$\lambda$} & \textbf{{\tiny86.4$^{0.3}$}} & {\tiny86.8$^{0.2}$} & {\tiny86.5$^{0.2}$} & {\tiny86.7$^{0.4}$} & \textbf{{\tiny84.5$^{0.4}$}} & \textbf{{\tiny85.0$^{0.1}$}} & \textbf{{\tiny85.2$^{0.2}$}} & \textbf{{\tiny85.5$^{0.2}$}} & \textbf{{\tiny77.4$^{1.4}$}} & \textbf{{\tiny79.0$^{0.8}$}} & \textbf{{\tiny79.7$^{0.9}$}} & \textbf{{\tiny79.1$^{0.8}$}} & \textbf{{\tiny61.6$^{0.5}$}} & \textbf{{\tiny68.8$^{0.5}$}} & \textbf{{\tiny71.6$^{0.5}$}} & \textbf{{\tiny74.0$^{0.7}$}}\\
{\tiny DPSH+HSWD} & {\tiny86.3$^{0.5}$} & \textbf{{\tiny86.9$^{0.3}$}} & \textbf{{\tiny86.9$^{0.2}$}} & {\tiny86.5$^{0.4}$} & {\tiny83.3$^{0.1}$} & {\tiny83.7$^{0.2}$} & {\tiny83.9$^{0.3}$} & {\tiny84.1$^{0.3}$} & {\tiny68.1$^{2.0}$} & {\tiny70.3$^{1.5}$} & {\tiny70.5$^{1.0}$} & {\tiny72.8$^{1.4}$} & {\tiny56.0$^{0.7}$} & {\tiny63.9$^{0.9}$} & {\tiny66.5$^{0.7}$} & {\tiny68.1$^{0.8}$}\\
{\tiny DPSH+H²Q($L_2$)} & {\tiny84.2$^{1.8}$} & {\tiny86.2$^{0.4}$} & {\tiny86.8$^{0.6}$} & {\tiny86.8$^{0.4}$} & {\tiny83.4$^{0.2}$} & {\tiny83.9$^{0.2}$} & {\tiny84.1$^{0.1}$} & {\tiny83.9$^{0.2}$} & {\tiny68.2$^{1.5}$} & {\tiny71.7$^{1.2}$} & {\tiny72.7$^{1.1}$} & {\tiny73.7$^{0.9}$} & {\tiny49.1$^{1.4}$} & {\tiny59.9$^{0.2}$} & {\tiny64.1$^{1.0}$} & {\tiny66.1$^{0.5}$}\\
{\tiny DPSH+H²Q($L_1$)} & {\tiny84.0$^{2.0}$} & {\tiny86.2$^{0.4}$} & {\tiny86.8$^{0.6}$} & {\tiny86.7$^{0.5}$} & {\tiny83.2$^{0.4}$} & {\tiny83.8$^{0.2}$} & {\tiny84.0$^{0.1}$} & {\tiny83.7$^{0.2}$} & {\tiny68.1$^{1.5}$} & {\tiny71.7$^{1.4}$} & {\tiny72.6$^{1.0}$} & {\tiny73.6$^{0.8}$} & {\tiny49.2$^{1.3}$} & {\tiny59.6$^{0.2}$} & {\tiny63.9$^{1.1}$} & {\tiny65.8$^{0.6}$}\\
{\tiny DPSH+H²Q(min)} & {\tiny84.1$^{1.8}$} & {\tiny86.2$^{0.3}$} & {\tiny86.8$^{0.6}$} & {\tiny86.7$^{0.4}$} & {\tiny83.4$^{0.2}$} & {\tiny83.8$^{0.2}$} & {\tiny84.0$^{0.1}$} & {\tiny83.8$^{0.3}$} & {\tiny68.0$^{1.4}$} & {\tiny72.0$^{1.3}$} & {\tiny72.6$^{1.1}$} & {\tiny73.7$^{0.9}$} & {\tiny49.0$^{1.1}$} & {\tiny59.1$^{0.6}$} & {\tiny63.7$^{1.1}$} & {\tiny66.1$^{0.7}$}\\
{\tiny DPSH+H²Q(bit)} & {\tiny84.1$^{1.9}$} & {\tiny86.3$^{0.3}$} & {\tiny86.8$^{0.5}$} & \textbf{{\tiny86.8$^{0.5}$}} & {\tiny83.4$^{0.2}$} & {\tiny83.9$^{0.2}$} & {\tiny84.1$^{0.1}$} & {\tiny83.8$^{0.3}$} & {\tiny68.0$^{1.6}$} & {\tiny71.8$^{1.2}$} & {\tiny72.4$^{1.1}$} & {\tiny73.6$^{0.9}$} & {\tiny48.7$^{0.7}$} & {\tiny59.2$^{0.7}$} & {\tiny63.8$^{1.0}$} & {\tiny66.0$^{0.6}$}\\[.2em]
{\tiny WGL.($\lambda=0$)} & {\tiny83.6$^{0.9}$} & {\tiny81.9$^{1.1}$} & {\tiny80.9$^{0.9}$} & {\tiny80.6$^{0.9}$} & {\tiny83.5$^{0.4}$} & {\tiny82.9$^{0.4}$} & {\tiny82.6$^{0.3}$} & {\tiny82.4$^{0.2}$} & {\tiny77.2$^{0.5}$} & {\tiny73.8$^{3.5}$} & {\tiny74.9$^{1.9}$} & {\tiny75.5$^{1.2}$} & {\tiny76.2$^{1.8}$} & {\tiny75.1$^{1.0}$} & {\tiny75.4$^{0.3}$} & {\tiny77.0$^{0.5}$}\\
{\tiny WGL.+ITQ} & {\color{red}{\tiny83.6$^{0.8}$}} & {\color{red}{\tiny81.9$^{1.3}$}} & {\color{red}{\tiny80.8$^{1.3}$}} & {\tiny81.2$^{0.5}$} & {\color{red}{\tiny83.5$^{0.4}$}} & {\tiny82.9$^{0.1}$} & {\color{red}{\tiny82.6$^{0.1}$}} & {\tiny82.7$^{0.1}$} & {\color{red}{\tiny77.1$^{0.4}$}} & {\tiny73.9$^{3.6}$} & {\tiny75.3$^{1.8}$} & {\tiny75.9$^{1.1}$} & {\tiny76.7$^{1.6}$} & {\color{red}{\tiny75.1$^{0.6}$}} & {\tiny75.9$^{0.5}$} & {\color{red}{\tiny77.0$^{0.4}$}}\\
{\tiny WGL.+$\lambda$} & {\tiny84.5$^{0.2}$} & {\tiny83.6$^{0.6}$} & \textbf{{\tiny84.2$^{0.2}$}} & {\tiny83.3$^{1.4}$} & {\tiny83.8$^{0.3}$} & {\tiny83.6$^{0.2}$} & {\tiny83.4$^{0.1}$} & {\tiny83.0$^{0.2}$} & {\tiny78.1$^{0.9}$} & {\tiny75.9$^{2.3}$} & {\tiny76.9$^{0.9}$} & {\tiny76.6$^{3.1}$} & {\tiny79.6$^{1.0}$} & \textbf{{\tiny79.4$^{0.6}$}} & \textbf{{\tiny79.0$^{0.8}$}} & {\tiny78.2$^{1.2}$}\\
{\tiny WGL.+HSWD} & {\tiny85.1$^{0.5}$} & {\tiny82.4$^{0.6}$} & {\color{red}{\tiny80.2$^{0.7}$}} & {\tiny80.8$^{0.9}$} & {\color{red}{\tiny83.0$^{0.3}$}} & {\color{red}{\tiny81.8$^{0.5}$}} & {\color{red}{\tiny81.1$^{0.4}$}} & {\color{red}{\tiny80.3$^{0.7}$}} & {\color{red}{\tiny75.9$^{0.6}$}} & \textbf{{\tiny79.6$^{0.4}$}} & \textbf{{\tiny78.0$^{0.6}$}} & {\tiny75.7$^{2.2}$} & \textbf{{\tiny79.7$^{0.5}$}} & {\tiny77.4$^{0.4}$} & {\color{red}{\tiny73.1$^{0.6}$}} & {\color{red}{\tiny69.2$^{1.6}$}}\\
{\tiny WGL.+H²Q($L_2$)} & {\tiny85.6$^{0.6}$} & {\tiny84.3$^{0.9}$} & {\tiny83.1$^{0.8}$} & {\tiny83.3$^{0.3}$} & {\tiny84.4$^{0.2}$} & {\tiny83.9$^{0.2}$} & {\tiny83.4$^{0.1}$} & \textbf{{\tiny83.4$^{0.2}$}} & \textbf{{\tiny79.4$^{0.4}$}} & {\tiny76.0$^{3.3}$} & {\tiny77.1$^{1.7}$} & \textbf{{\tiny77.6$^{1.0}$}} & {\tiny79.6$^{1.3}$} & {\tiny78.5$^{0.5}$} & {\tiny78.4$^{0.2}$} & \textbf{{\tiny79.2$^{0.6}$}}\\
{\tiny WGL.+H²Q($L_1$)} & {\tiny85.4$^{0.5}$} & {\tiny84.4$^{0.9}$} & {\tiny83.0$^{0.6}$} & {\tiny83.4$^{0.4}$} & {\tiny84.3$^{0.3}$} & {\tiny84.0$^{0.2}$} & \textbf{{\tiny83.5$^{0.1}$}} & {\tiny83.3$^{0.3}$} & {\tiny79.3$^{0.4}$} & {\tiny76.0$^{3.3}$} & {\tiny77.1$^{1.5}$} & {\tiny77.5$^{1.1}$} & {\tiny79.5$^{1.3}$} & {\tiny78.1$^{0.7}$} & {\tiny78.3$^{0.4}$} & {\tiny78.9$^{0.3}$}\\
{\tiny WGL.+H²Q(min)} & {\tiny85.7$^{0.4}$} & {\tiny84.4$^{0.9}$} & {\tiny83.3$^{0.7}$} & \textbf{{\tiny83.6$^{0.3}$}} & {\tiny84.3$^{0.2}$} & \textbf{{\tiny84.0$^{0.1}$}} & {\tiny83.4$^{0.1}$} & {\tiny83.4$^{0.2}$} & {\tiny79.3$^{0.4}$} & {\tiny76.0$^{3.3}$} & {\tiny77.2$^{1.5}$} & {\tiny77.6$^{1.1}$} & {\tiny79.6$^{1.3}$} & {\tiny77.8$^{0.8}$} & {\tiny78.0$^{0.2}$} & {\tiny78.9$^{0.4}$}\\
{\tiny WGL.+H²Q(bit)} & \textbf{{\tiny85.7$^{0.3}$}} & \textbf{{\tiny84.5$^{0.9}$}} & {\tiny83.1$^{0.7}$} & {\tiny83.6$^{0.4}$} & \textbf{{\tiny84.4$^{0.2}$}} & {\tiny84.0$^{0.1}$} & {\tiny83.3$^{0.1}$} & {\tiny83.4$^{0.3}$} & {\tiny79.3$^{0.4}$} & {\tiny75.9$^{3.4}$} & {\tiny77.0$^{1.7}$} & {\tiny77.5$^{1.1}$} & {\tiny79.2$^{1.3}$} & {\tiny77.4$^{0.7}$} & {\tiny77.5$^{0.3}$} & {\tiny78.5$^{0.3}$}\\[.2em]
{\tiny HyP²($\lambda=0$)} & {\tiny85.0$^{0.8}$} & {\tiny85.2$^{0.2}$} & {\tiny85.6$^{0.1}$} & {\tiny85.7$^{0.3}$} & {\tiny85.1$^{0.2}$} & {\tiny85.5$^{0.2}$} & {\tiny85.9$^{0.2}$} & {\tiny85.9$^{0.2}$} & {\tiny79.7$^{1.0}$} & {\tiny82.0$^{0.1}$} & {\tiny82.6$^{0.6}$} & {\tiny82.2$^{0.1}$} & {\tiny75.7$^{0.5}$} & {\tiny77.5$^{0.7}$} & {\tiny79.0$^{0.7}$} & {\tiny78.8$^{0.3}$}\\
{\tiny HyP²+ITQ} & {\color{red}{\tiny84.7$^{0.5}$}} & {\color{red}{\tiny85.0$^{0.3}$}} & {\color{red}{\tiny85.6$^{0.5}$}} & {\color{red}{\tiny85.5$^{0.5}$}} & {\color{red}{\tiny85.1$^{0.3}$}} & {\tiny85.6$^{0.1}$} & {\color{red}{\tiny85.8$^{0.1}$}} & {\color{red}{\tiny85.9$^{0.2}$}} & {\color{red}{\tiny79.6$^{1.0}$}} & {\color{red}{\tiny81.9$^{0.2}$}} & {\color{red}{\tiny82.6$^{0.6}$}} & {\color{red}{\tiny82.1$^{0.3}$}} & {\color{red}{\tiny75.1$^{0.3}$}} & {\color{red}{\tiny77.0$^{0.6}$}} & {\color{red}{\tiny78.4$^{0.4}$}} & {\color{red}{\tiny78.8$^{0.5}$}}\\
{\tiny HyP²+$\lambda$} & \textbf{{\tiny86.1$^{0.6}$}} & {\tiny86.1$^{0.3}$} & \textbf{{\tiny86.6$^{0.1}$}} & \textbf{{\tiny86.5$^{0.5}$}} & {\tiny85.3$^{0.1}$} & {\tiny85.6$^{0.1}$} & {\color{red}{\tiny85.6$^{0.2}$}} & {\color{red}{\tiny85.4$^{0.1}$}} & {\tiny81.3$^{0.4}$} & {\color{red}{\tiny81.2$^{0.5}$}} & {\color{red}{\tiny79.8$^{0.5}$}} & {\color{red}{\tiny78.2$^{0.2}$}} & \textbf{{\tiny79.3$^{0.8}$}} & \textbf{{\tiny82.0$^{0.4}$}} & \textbf{{\tiny81.6$^{0.6}$}} & {\tiny80.9$^{0.1}$}\\
{\tiny HyP²+HSWD} & {\tiny85.8$^{0.8}$} & {\tiny85.6$^{0.3}$} & {\tiny85.8$^{0.3}$} & {\tiny86.0$^{0.5}$} & {\tiny85.1$^{0.3}$} & {\color{red}{\tiny85.4$^{0.2}$}} & {\color{red}{\tiny85.7$^{0.1}$}} & {\color{red}{\tiny85.7$^{0.1}$}} & {\tiny80.1$^{0.5}$} & {\color{red}{\tiny81.8$^{0.1}$}} & {\color{red}{\tiny82.2$^{0.6}$}} & {\color{red}{\tiny81.6$^{0.2}$}} & {\tiny77.3$^{0.5}$} & {\tiny79.3$^{0.2}$} & {\tiny80.0$^{0.4}$} & {\tiny78.8$^{0.8}$}\\
{\tiny HyP²+H²Q($L_2$)} & {\tiny86.0$^{0.6}$} & {\tiny86.3$^{0.3}$} & {\tiny86.4$^{0.3}$} & {\tiny86.4$^{0.4}$} & \textbf{{\tiny85.7$^{0.3}$}} & \textbf{{\tiny86.0$^{0.1}$}} & {\tiny86.1$^{0.1}$} & {\tiny86.1$^{0.1}$} & {\tiny81.4$^{0.5}$} & {\tiny82.7$^{0.2}$} & \textbf{{\tiny83.1$^{0.1}$}} & {\tiny82.4$^{0.2}$} & {\tiny77.9$^{0.9}$} & {\tiny80.6$^{0.5}$} & {\tiny81.5$^{0.4}$} & \textbf{{\tiny81.5$^{0.2}$}}\\
{\tiny HyP²+H²Q($L_1$)} & {\tiny86.0$^{0.4}$} & {\tiny86.1$^{0.2}$} & {\tiny86.5$^{0.3}$} & {\tiny86.4$^{0.4}$} & {\tiny85.7$^{0.2}$} & {\tiny85.9$^{0.2}$} & {\tiny86.1$^{0.1}$} & {\tiny86.1$^{0.1}$} & {\tiny81.4$^{0.5}$} & {\tiny82.7$^{0.3}$} & {\tiny83.0$^{0.2}$} & \textbf{{\tiny82.4$^{0.2}$}} & {\tiny77.8$^{0.9}$} & {\tiny80.3$^{0.3}$} & {\tiny81.2$^{0.5}$} & {\tiny81.3$^{0.2}$}\\
{\tiny HyP²+H²Q(min)} & {\tiny85.9$^{0.4}$} & {\tiny86.3$^{0.3}$} & {\tiny86.4$^{0.3}$} & {\tiny86.3$^{0.3}$} & {\tiny85.7$^{0.2}$} & {\tiny86.0$^{0.1}$} & \textbf{{\tiny86.1$^{0.1}$}} & \textbf{{\tiny86.1$^{0.1}$}} & {\tiny81.4$^{0.5}$} & \textbf{{\tiny82.7$^{0.2}$}} & {\tiny83.1$^{0.1}$} & {\tiny82.4$^{0.1}$} & {\tiny77.6$^{0.9}$} & {\tiny80.2$^{0.4}$} & {\tiny81.1$^{0.4}$} & {\tiny81.3$^{0.2}$}\\
{\tiny HyP²+H²Q(bit)} & {\tiny86.0$^{0.5}$} & \textbf{{\tiny86.4$^{0.2}$}} & {\tiny86.3$^{0.3}$} & {\tiny86.3$^{0.3}$} & {\tiny85.7$^{0.2}$} & {\tiny86.0$^{0.1}$} & {\tiny86.1$^{0.1}$} & {\tiny86.1$^{0.0}$} & \textbf{{\tiny81.6$^{0.6}$}} & {\tiny82.7$^{0.2}$} & {\tiny83.0$^{0.1}$} & {\tiny82.4$^{0.2}$} & {\tiny77.3$^{0.6}$} & {\tiny79.8$^{0.4}$} & {\tiny80.9$^{0.5}$} & {\tiny81.4$^{0.2}$}\\[.2em]
\end{tabular}
\caption{\footnotesize{Full comparison of \texttt{mAP@k} before (level $\lambda = 0$) and after using each quantization strategy on VGG-16. Performance metrics in red indicate a decrease in performance after the quantization strategy and performance metrics in bold indicate which quantization strategy gave the best overall metric. The small superscript numbers indicate the standard deviation of the metrics.}}
\label{sm-tab:full_CNNF_vgg16_SM}
\end{table*}

\subsection{Comparison Between Quantization Strategies}
\label{sm-sec:comp_quant_strat}

Table \ref{sm-tab:full_CNNF_vgg16_SM} shows the full results depicted in Figure \ref{fig:other_quantizations} and Table \ref{sm-tab:full_CNNF_alexnet_SM} contains the analogous results for AlexNet. As discussed in Section \Cref{sec:other_quantizations} of the main text, H²Q stands out as the only strategy which never decreases performance, regardless of the choice of loss function. ITQ frequently decreases performance metrics, and is never the best overall strategy. ITQ has an average improvement of $0$ percentage points (p.p.) (after rounding) on both architectures. HWSD does not reduce the performance metrics as often as ITQ and it is the best overall strategy in $13$ cases in AlexNet and $16$ cases in VGG-16. Its average improvement is $1.6$ p.p. on AlexNet and $1.0$ p.p. on VGG-16. The use of a penalty term is the overall best strategy in $39$ cases on AlexNet and $34$ cases on VGG-16, but its average improvement is $1.2$ p.p. on AlexNet and $-0.3$ p.p. on VGG-16. Even though it makes substantiall improvements in many cases, it sometimes has a very negative impact on performance metrics, \eg ImageNet using CEL. At last, H²Q is not always the best overall method, but it wins in $44$ cases on AlexNet and $46$ cases on VGG-16, with the advantage of never decreasing performance, as mentioned earlier. Even if we consider only the $L_2$ loss, H²Q is the best overall in $43$ cases on AlexNet and $45$ cases on VGG-16 with an average improvement of $2.09$ p.p. and $1.83$ p.p. respectively.

\begin{figure}
    \centering
    \includegraphics[width=.5\linewidth]{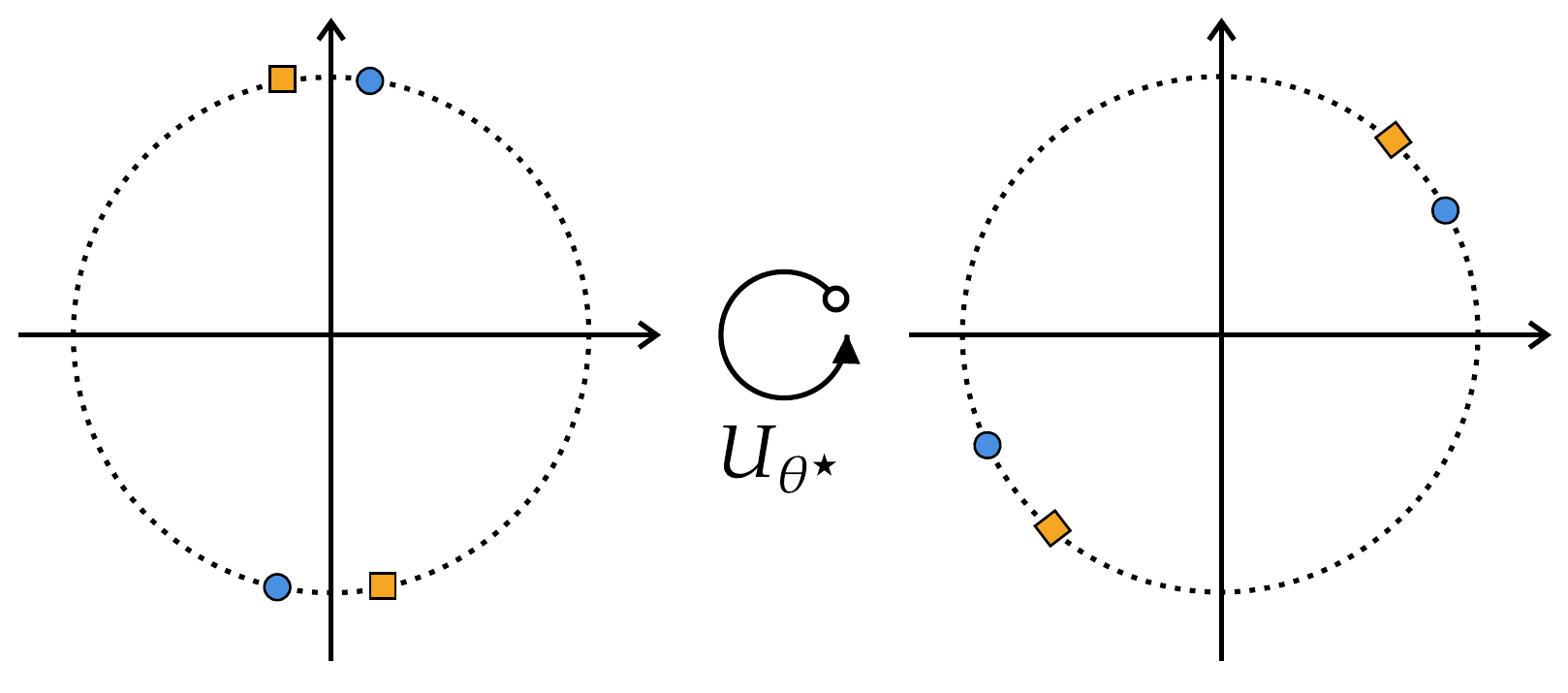}
    \caption{Example of a case where the unsupervised losses could decrease the $\texttt{mAP}$.}
    \label{sm-fig:bad_case}
\end{figure}

Moreover, the performances of all losses experimented on our ablation study, \ie, the $L_2$, $L_1$, min entry and bit var losses, is close. This points to the fact that, at least among unsupervised losses, the choice of the loss is not a relevant issue. This might not be the case for supervised losses, \ie, it may be the case that supervised losses can be used to find better rotations. A clear example of this is given in Figure \ref{sm-fig:bad_case}, where two classes are given (blue circles and orange rectangles), but the classes are poorly separated. In such a case our proposed quantization strategy would decrease the $\texttt{mAP}$ since it would put the points closer to the $y=x$ line, putting dissimilar points on the same hash. This undesirable behavior could be avoided using \HQ\ with supervised losses, but since our experiments have shown improvements in all cases (see Tables \ref{sm-tab:full_CNNF_alexnet_SM} and \ref{sm-tab:full_CNNF_vgg16_SM}), we conjecture that ill-posed geometries such as Figure \ref{sm-fig:bad_case} are rare in practice, which is simply to say that the similarity learning strategies available on the literature are capable of provide well-separeted embeddings.

\section{Proofs}
\label{sm:proofs}

We now give self-contained proofs of Theorems \ref{thm:inner_prod} and \ref{thm:householder_decomp}.

\subsection{Proof of \Cref{thm:inner_prod}}

\begin{proof}
    Let $f: \mathbb{R}^k \to \mathbb{R}^k$ be such that $\langle f(u), f(v) \rangle = \langle u, v \rangle$ for all $u,v \in \mathbb{R}^k$. Given $\lambda \in \mathbb{R}$ and $u, v \in \mathbb{R}^k$ we have that:
    \begin{equation}
        \begin{split}
            \| \lambda f(u) - f(\lambda u) \|^2 &= \lambda^2 \| f(u) \|^2 + \| f(\lambda u) \|^2 \\ 
            &- 2 \langle \lambda f(u), f(\lambda u) \rangle \\
            &= 2 \lambda^2 \| u \|^2 - 2 \lambda^2 \langle u, u \rangle \\
            &= 0.
        \end{split}
    \end{equation}
    Also, note that
    \begin{equation}
        \begin{split}
            \| f(u+v) - & f(u) - f(v) \|^2\\
            & = \| f(u + v) \|^2 + \| f(u) + f(v) \|^2 \\
            & \,\,\,\,\,\, - 2 \langle f(u + v), f(u) + f(v) \rangle \\
            & = \|u + v \|^2 + \| u \|^2 + \| v \|^2 + 2 \langle u, v \rangle  \\
            & \,\,\,\,\,\, - 2 \left( \langle u + v, u \rangle + \langle u + v, v \rangle \right) \\
            & = 2 \| u + v \|^2 - 2 \| u + v \|^2 \\
            & = 0.
        \end{split}
    \end{equation}
    This shows that $f$ is linear. The fact that it is orthogonal follows by the fact the it preserves inner products.
\end{proof}

\subsection{Proof of \Cref{thm:householder_decomp}}

For \Cref{thm:householder_decomp} we follow \cite{uhlig2001constructive} and \cite{golub1996matrix}:

\begin{proof}

Given $x \in \mathbb{R}^k \setminus \{ 0 \}$, if we take
\begin{equation}
    H = I_k - 2 \frac{v v^{T}}{\| v \|_2^2}
\end{equation}
with $v = x - \|x\|_2 e_1$, where $e_1 = (1, 0, \ \dots , 0)^T$, we have
\begin{equation}
    Hx = \|x\|_2 e_1.
\end{equation}

It follows from this that we can transform any $k \times k$ invertible matrix $A$ into an upper-triangular matrix using successive multiplications by Householder matrices. In particular, given $U$ an orthogonal matrix there exists $H_1, \dots, H_{k-1}$ Householder matrices such that:

\begin{equation}
\label{eq:upp_triang}
    H_{k-1} \cdots H_1 U = R,
\end{equation}
where $R$ is upper-triangular. Since each term on the product on the left hand side of \eqref{eq:upp_triang} is an orthogonal matrix, it follows that $R$ must also be orthogonal, thus $R^T R = I_k$. Letting $r_i$ denote the $i$-th column of $R$, this implies that $\| r_1 \|^2 = r^2_{11} = 1$ so $r_{11} = \pm 1$. Also, $\langle r_j, r_1 \rangle = r_{j1} r_{11} = 0$ for $j = 2, \dots, k$ so $r_{j1} = 0$ for $j = 2, \dots, k$ since $r_{11} \neq 0$. Repeating this reasoning for the second column and so on we have that $R$ must be diagonal with entries $\pm 1$. We will naturally have that $r_{ii} = 1$ for $i=1, \dots, k-1$ by the construction of $H_i$. If $r_{kk} = -1$, we do an additional multiplication by $H_k = I_k - 2e^{T}_k e_k$. Taking $H_k = I_k$ in case $r_{kk} = 1$, it follows that we can write:
\begin{equation*}
    I_k = H_k \dots H_1 U.
\end{equation*}
Since each $H_i$ is symmetric and orthogonal, it follows that:

\begin{equation*}
    H_1 \dots H_k = U.
\end{equation*}

The fact that each product of Householder matrices is orthogonal follows directly from the fact that each Householder matrix is orthogonal and $O(k)$ is closed under multiplications. \end{proof}

\end{document}